\documentclass[journal]{IEEEtran}
\usepackage{amsmath,amsfonts}

\usepackage{array}
\usepackage[caption=false,font=normalsize,labelfont=sf,textfont=sf]{subfig}
\usepackage{textcomp}
\usepackage{stfloats}
\usepackage{url}
\usepackage{verbatim}
\usepackage{graphicx}
\usepackage{amsmath, bm}
\usepackage{multirow}
\usepackage{bm}
\usepackage{caption}
\newtheorem{mydef}{Definition}

\usepackage{makecell}
\usepackage{booktabs} 
\usepackage{threeparttable} 
\renewcommand{\arraystretch}{0.8} 
\usepackage{etoolbox}
\makeatletter
\patchcmd{\@makecaption}
  {\scshape}
  {}
  {}
  {}
\makeatletter
\patchcmd{\@makecaption}
  {\\}
  {.\ }
  {}
  {}
\makeatother

\usepackage{lineno,hyperref}
\usepackage{hyperref}
\usepackage{cite}
\usepackage{mathrsfs}
\hypersetup{hidelinks}
\hypersetup{
citecolor=blue,
colorlinks=true,
linkcolor=blue,
urlcolor=blue
}
\usepackage{graphicx}
\usepackage{graphics}
\usepackage{subfloat}
\usepackage{float}
\graphicspath{{pic/}} 
\usepackage{cleveref}

\usepackage{float}
\usepackage[ruled,linesnumbered]{algorithm2e}
\SetKwRepeat{Do}{do}{while}%

\usepackage{graphicx}

\usepackage[export]{adjustbox}

\begin{document}

\title{GBG++: A Fast and Stable Granular Ball Generation Method for Classification}

\author{Qin Xie,
    Qinghua Zhang*,
    Shuyin Xia,
    Fan Zhao,
    Chengying Wu,
	Guoyin Wang,
    Weiping Ding,
	
}

\maketitle   

\begin{abstract}
Granular ball computing (GBC), as an efficient, robust, and scalable learning method, has become a popular research topic of granular computing. GBC includes two stages: granular ball generation (GBG) and multi-granularity learning based on the granular ball (GB). However, the stability and efficiency of existing GBG methods need to be further improved due to their strong dependence on $k$-means or $k$-division. In addition, GB-based classifiers only unilaterally consider the GB's geometric characteristics to construct classification rules, but the GB's quality is ignored. Therefore, in this paper, based on the attention mechanism, a fast and stable GBG (GBG++) method is proposed first. Specifically, the proposed GBG++ method only needs to calculate the distances from the data-driven center to the undivided samples when splitting each GB instead of randomly selecting the center and calculating the distances between it and all samples. Moreover, an outlier detection method is introduced to identify local outliers. Consequently, the GBG++ method can significantly improve effectiveness, robustness, and efficiency while being absolutely stable. Second, considering the influence of the sample size within the GB on the GB's quality, based on the GBG++ method, an improved GB-based $k$-nearest neighbors algorithm (GB$k$NN++) is presented, which can reduce misclassification at the class boundary. Finally, the experimental results indicate that the proposed method outperforms several existing GB-based classifiers and classical machine learning classifiers on $24$ public benchmark datasets. The implementation code of experiments is available at https://github.com/CherylTse/GBG-plusplus.
\end{abstract}


\section{Introduction}
\label{sec:Introduction}
\IEEEPARstart{D}{ata} mining encounters new challenges in the era of information technology. Specifically, the effective and efficient knowledge discovery for large-scale, high-dimensional, and sparse data is the focus of current research\cite{55,56,34}. As an important part of artificial intelligence (AI), cognitive computing is an emerging field involving the collaborative integration of cognitive science, data science, and a series of technologies \cite{1}. The purpose of cognitive computing is to establish a calculation mechanism that simulates the functions of the human brain for knowledge discovery from complex data. In cognitive science, due to the bottleneck of information processing, humans can only process part of the information when faced with complex scenarios. Additionally, Chen \cite{7} highlights that human cognition follows the rule of `global topology precedence'. For instance, the visual system is first susceptible to the global topological characteristics and then will focus on the local characteristics. In essence, this is a multi-granularity cognitive mechanism through which the perspective of human beings to obtain information is shifted from the global to the local.
Based on this cognitive mechanism, granular computing (GrC) \cite{8}, proposed by Zadeh, is a series of methods of machine intelligence and cognitive computing methodologies that use the information granule (IG) as the computing unit, which can significantly improve the efficiency for knowledge discovery. Rodriguez \cite{9} points out that GrC is an effective data mining method for processing massive data. In addition, compared with the existing deep learning method\cite{28,53,54}, the GrC has the advantages of interpretability, high computational efficiency, and fewer hyperparameters.
After decades of development, the GrC has achieved significant results. Moreover, scholars have put forward some theories such as fuzzy sets \cite{10,11,49}, rough sets \cite{12}, quotient space theory \cite{13}, and three-way decisions \cite{14,15}.
Besides, some extensions\cite{47,23,20,22} have been proposed based on the aforementioned theories, which are mainly used to provide new explanations for previous methods.

\begin{figure}[htbp]
\vspace{-0.1in}
	\centering
	\begin{minipage}[t]{0.7\linewidth}
		\centering
		\includegraphics[height=0.7in,width=2.3in]{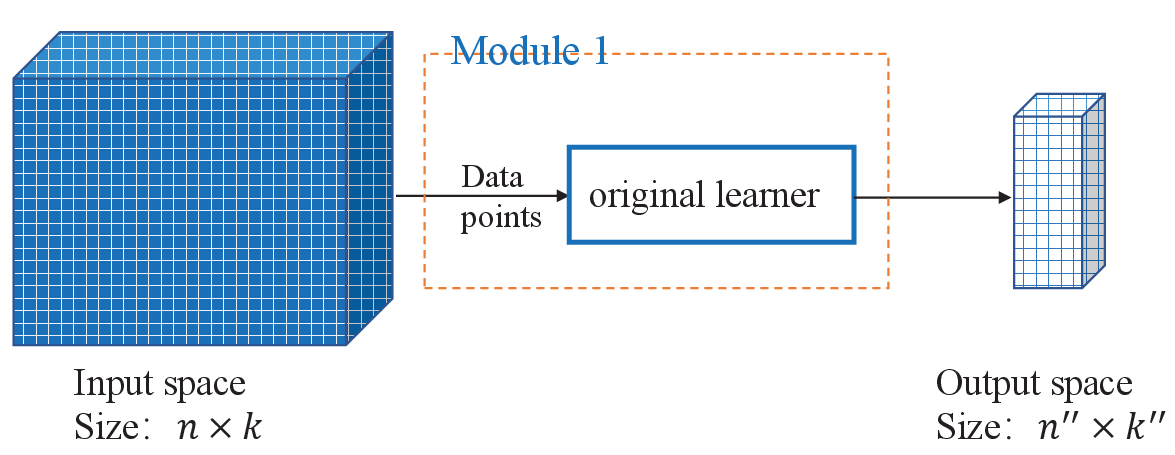}
		\small (a) Traditional machine learning process.
	\end{minipage}%
	\vspace{0.2cm}
	\begin{minipage}[t]{0.7\linewidth}
		\centering
		\includegraphics[height=0.7in,width=2.5in]{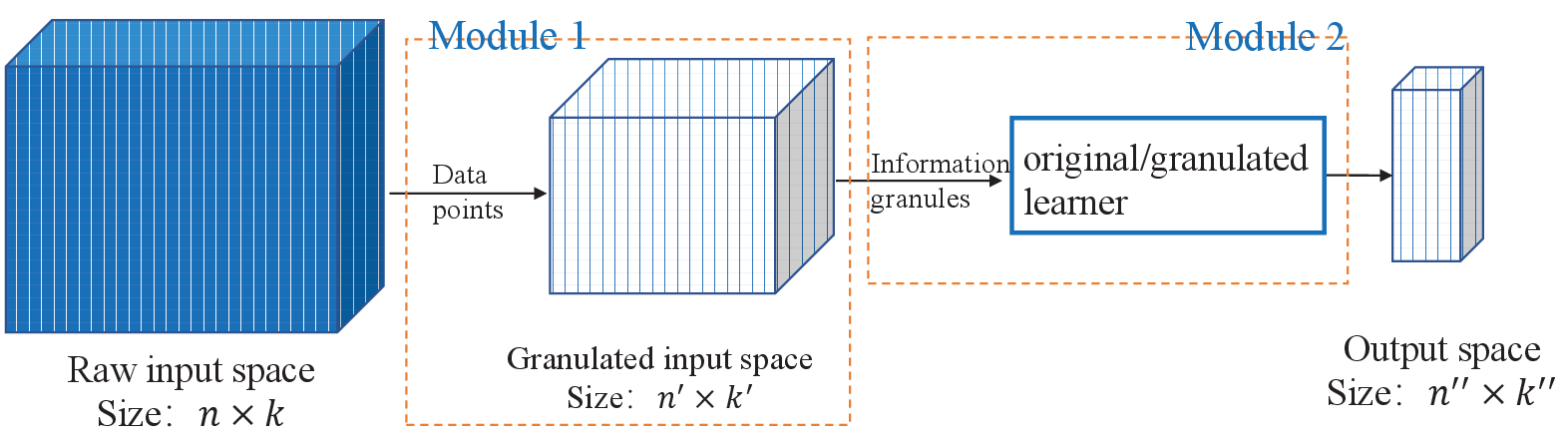}
		\small (b) Multi-granularity learning process.
	\end{minipage}%
	\caption {Comparison of multi-granularity learning and traditional machine learning: multi-granularity learning involves an additional data granulation module.}
	\vspace{0.1in}
	\label{fig1}
\vspace{-0.5cm}
\end{figure}

In the field of GrC, previous research has indicated that the larger the granularity, the higher the algorithm execution efficiency, and the better the noise robustness, but the worse the effectiveness. The schematic diagram of the learning process of traditional machine learning is shown in Fig. \ref{fig1}(a), but the execution efficiency of most of the algorithms\cite{21,28,50,52,53} needs to be improved. From the perspective of GrC, one of the reasons is that the granularity of the input space is too fine, which is the most fine-grained data space whose computing unit is the data point. However, for different application scenarios, an appropriate granularity can ensure both effectiveness and efficiency, which is called multi-granularity learning (MGL), shown in Fig. \ref{fig1}(b). In general, MGL is a two-stage learning model that includes data granulation (Module 1 in Fig. \ref{fig1}(b)) and granule-based machine learning (Module 2 in Fig. \ref{fig1}(b)). Data granulation is to decompose the complex dataset into IGs based on a given strategy. As shown in Fig. \ref{fig1}, compared to traditional machine learning, MGL involves an additional module that is used for granulating the input data space into multiple subspaces to obtain a granulated data space whose dimension is generally smaller. Thus, the IGs are fed into the learner rather than the data points, as shown in Module 2 in Fig. \ref{fig1}(b). However, the mainstream binary-relation-based data granulation methods \cite{18,19} are time-consuming. Up to now, there are some representative MGL methods. For instance, Pedrycz et al.\cite{24} propose a new computing architecture that focuses on processing IGs, in which the input and output layers can be granulated using the rough set or fuzzy set methods. And, a new architecture of a granular neural network is presented\cite{25}, in which the IGs in the multidimensional input space are formed using feature selection. Besides, a long-term prediction approach based on the back-propagation neural network using IGs as input is given\cite{26}. However, in these methods, GrC is used for feature selection or representation without optimizing the structure of the neural network. Therefore, Ding et al. \cite{27} present a weighted linear loss multiple birth support vector machine (SVM) by dividing the dataset into several granules and constructing a set of sub-classifiers. Furthermore, using granulated layers to recognize different object sizes and shapes, a granulated region-based convolutional neural network is proposed\cite{48}. However, these listed methods sacrifice efficiency. Hence, considering the scenarios of large-scale data mining, the MGL methods should be highly efficient, scalable, and robust.

Given this, Xia et al.\cite{29} propose GBC, an efficient, scalable, and robust MGL method. Fig. \ref{fig1}(b) can describe the structure of GBC when the granulated input space is composed of GBs using the GBG method. In the stage of GBG, the dataset is divided into multiple subsets, and the GB is constructed on each subset using the ball, which can reduce the dimensions of the dataset because the ball can be described using only two parameters. Compared with the traditional data granulation method based on binary relation, the time complexity of the GBG method, which is linear, has been dramatically improved. As for the stage of GB-based machine learning, the GBs are used to replace the data points, and the structure of the learner can be optimized using the idea of multi-granularity. After several years of development, GBC has been extended to some theories and application scenarios. For instance, the efficiency of the GB-based $k$NN \cite{29} is hundreds of times higher than that of the existing $k$NN\cite{33,35,36}, particularly for large-scale datasets. In addition, the GBC is introduced into the $k$-means, and a simple and fast $k$-means clustering method called ball $k$-means is developed \cite{30}. Moreover, a general sampling method called GB sampling is presented that uses adaptively constructed GBs to cover the data space, and the samples on the GBs constitute the sampled dataset\cite{31}.
In addition, an acceleration GBG method \cite{32} that replaces $k$-means with $k$-division is proposed, which can significantly improve the efficiency of GBC while ensuring an accuracy similar to that of the existing GBC. As discussed, although much effort has been dedicated to GBC, the noted methods mainly suffer from limitations and challenges as follows. 1) Existing GBG methods rely on clustering methods, such as $k$-means and $k$-division, which need to randomly select the cluster centers; that is, the existing methods are not stable. 2) Since $k$-means and $k$-division need to calculate the distances from each center to all samples, the existing GBG methods are still time-consuming. 3) Existing GBG methods consider it reasonable that the GB contains only a sample; however, most of these samples are outliers. 4) Existing GB-based classifiers only unilaterally consider the center and radius of GB to construct the classification rule, while the quality of the GB is ignored.

The attention mechanism (AM) originates from the nuclear regression problem proposed by Naradaya Watson \cite{2}, which can improve the effectiveness and efficiency of machine learning. And the AM can be divided into two classes: soft attention\cite{3,4} and hard attention\cite{5,6}. By introducing AM, a learner can prioritize the most crucial parts when dealing with large amounts of data, thereby reducing computation and storage overhead. The core idea of hard attention is to select the most relevant element in a given input sequence.
Therefore, inspired by hard attention in AM, a fast and stable GBG (GBG++) method is proposed in this paper. Further, based on the GBG++ method, a GB-based $k$NN classifier (GB$k$NN++) is given. The main contributions of this paper are as follows.

1) The AM is introduced into the method of splitting the GB to make the construction process of the GB attention-based data-driven. In addition, the proposed GB splitting method only needs to calculate the distances between the center and undivided samples. Consequently, the GBG++ method can accelerate the existing GBG method several times to dozens of times while maintaining its stability and effectiveness.

2) The proposed outlier detection method is capable of identifying local outliers while splitting the GB, thereby tackling the problem of some outliers being incorrectly constructed as valid GBs. In addition, the outlier detection method can help reduce misclassification at the class boundary as well as the number of GBs, which can further improve the effectiveness and efficiency of classification tasks.

3) The framework of the GB-based classifier is redescribed. Moreover, the GB$k$NN++ is proposed to tackle the problem of ignoring the quality of GB when constructing the classification rule.

The remainder of this paper is organized as follows. Section \ref{sec:RW} reviews related works on the original GBG method, an acceleration GBG method, and the original GB$k$NN. In Section \ref{sec:GBG++}, the GBG++ method is proposed and described in detail. Section \ref{sec:GB$k$NN++} introduces the GB$k$NN++. The performance of the proposed method is demonstrated on some benchmark datasets in Section \ref{sec:Experiments}. Finally, the conclusions and further work are presented in Section \ref{sec:Conclusion}.

\section{Related Work}
\label{sec:RW}

\noindent To make the paper more concise, in the subsequent content, let $D(D=\{(\bm{x_1},y_1),(\bm{x_2},y_2),...,(\bm{x_n},y_n)\})$ be a dataset. Moreover, $\bm{X}(\bm{X}=[\bm{x_1},\bm{x_2},...,\bm{x_n}],\bm{x_i} \in {\mathbb{R}}^q,i=1,2,...,n)$ represents the feature value matrix of $D$ with $q$ features, and $\bm{Y}(\bm{Y}=[y_1,y_2,...,y_n],y_i\in \mathbb{R})$ is the corresponding label vector. In addition, $G(G=\{gb_1,gb_2,...,gb_m\})$ is a set of GBs generated on $D$. Samples with the same label are called homogeneous samples; otherwise, they are called heterogeneous samples. The set composed of the majority of samples in $D$ is denoted as the majority class of $D$.

\vspace{-0.3cm}

\subsection{Original GBG Method}
\label{subsec:Ori-GBG}

\begin{figure}[htbp]
\vspace{-0.1in}
\centering
\includegraphics[height=0.8in,width=2.7in]{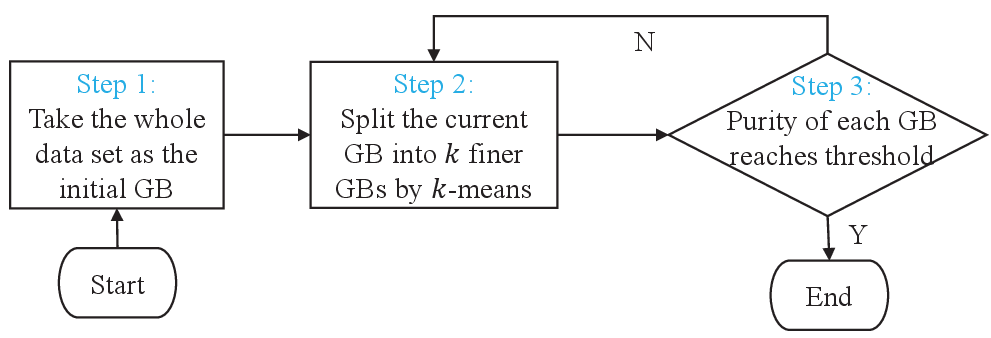}
\caption {Workflow of original GBG method.}
\label{fig2}
\vspace{-0.2in}
\end{figure}

The core idea of the GBG method is to cover a dataset with a set of balls, where each ball is called a GB which is indeed an IG. Additionally, the granulation process using the original GBG method can be briefly described as these steps shown in Fig. \ref{fig2}. Step 1 uses the whole training dataset as the initial GB. For Step 2, $k$-means is employed to split the GB into $k$ finer GBs whose centers and radii are determined based on Definition \ref{mydef1}. Notably, Xia\cite{29} points out that considering both the computation time and the quality of GB, the value of $k$ can be set to 2.

\begin{mydef}\label{mydef1} \cite{29} Given a dataset $D$. Suppose $G$ be a set of GBs generated on $D$. For $\forall gb_i\in G$, it is generated on $D_i (D_i \subseteq D)$, and the center $\bm{c_i}$ and radius $r_i$ of $gb_i$ are respectively defined as follows,
\begin{equation}\label{eq1}
\bm{c_i}=\frac{1}{|D_i|}\sum_{(\bm{x},y)\in D_i}\bm{x},
\end{equation}
\begin{equation}\label{eq2}
r_i=\frac{1}{|D_i|}\sum_{(\bm{x},y) \in D_i} \bigtriangleup(\bm{x},\bm{c_i}),
\end{equation}
where $|\bullet|$ represents the cardinality of set $\bullet$, and $\bigtriangleup(\cdot)$ denotes the distance function. And, without losing generality, all distances in this paper refer to Euclidean distances.
\end{mydef}

The quality of the GB is measured using the purity defined in Definition \ref{mydef2}. The closer the purity is to $1$, the better the coverage effect of the GBs on the original dataset. Meanwhile, the granulation stopping condition is controlled using a specified purity threshold. As shown in Step 3 of Fig. \ref{fig2}, for each generated GB, judge whether the purity of the GB reaches the threshold. If not, the GB returns to Step 2.

\begin{mydef}\label{mydef2} \cite{29} Given a dataset $D$. Suppose $G$ be a set of GBs generated on $D$. For $\forall gb_i \in G$, it's label and purity are defined respectively as follows,
\begin{equation}\label{eq3}
l_i={ \underset{l_j \in L_i}{{\arg\max} \,} |\{(\bm{x},y)\in gb_i|y=l_j\}|},
\end{equation}
\begin{equation}\label{eq4}
p_i=\frac{|\{(\bm{x},y)\in gb_i|y=l_i\}|}{|gb_i|},
\end{equation}
\end{mydef}
where $L_i$ represents the set of label category of samples contained in $gb_i$.

According to Definition \ref{mydef2}, since the label of the GB is determined by the majority of samples contained in the GB, the label would not generally be affected by outliers. Therefore, the original GBG method is robust.

\vspace{-0.3cm}

\subsection{An Acceleration GBG Method}
\label{subsec:ACC-GBG}

\begin{figure}[htbp]
\vspace{-0.1in}
\centering
\includegraphics[height=0.8in,width=3.2in]{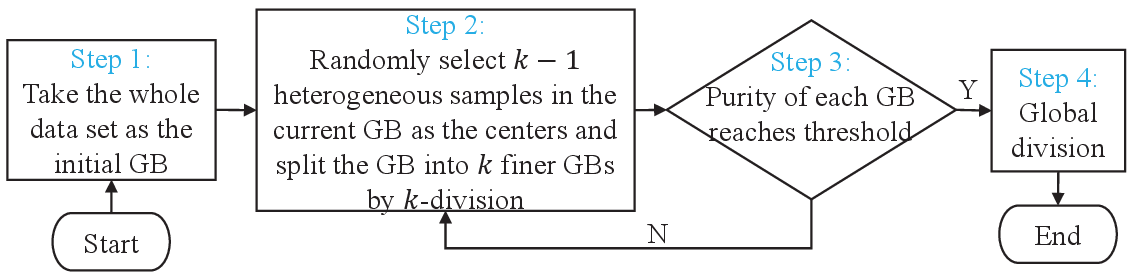}
\caption {Workflow of acceleration GBG method.}
\label{fig3}
\vspace{-0.2in}
\end{figure}

The core idea of the acceleration GBG method\cite{32} is to replace the $k$-means used in the original GBG method with the $k$-division. The granulation process of the acceleration GBG method can be briefly described in several steps, as shown in Fig. \ref{fig3}. Steps 1 and 3 are the same as those in the original GBG method. In Step 2, $k$-division splits the GB into $k$ finer GBs with heterogeneous centers. In addition, $k$ denotes the number of label categories of the samples contained in the GB. Besides, the centers selected by the $k$-division are only the division centers, not the centroids of balls calculated using Definition \ref{mydef1}. Thus, as shown in Step 4 of Fig. \ref{fig3}, the acceleration GBG method performs a global $m$-division on the whole training dataset so that the division center can be close to the center of the corresponding GB, in which $m$ represents the number of all generated GBs.

\subsection{Original GB$k$NN}
\label{subsec:Ori-GB$k$NN}

The core idea of GB$k$NN is that the label of each queried sample is determined by the nearest GB whose label is determined by the majority of samples within the GB. The distance between the queried sample and the GB is defined as follows.

\begin{mydef}\label{mydef3}\cite{29} Given a dataset $D$ and a queried sample $\bm{x}$. Suppose $G$ be a set of GBs generated on $D$. For $\forall gb_i \in G $ with center $\bm{c_i}$ and radius $r_i$, the distance between $\bm{x}$ and $gb_i$ is defined as follows,
\begin{equation}\label{eq6}
dis(\bm{x},gb_i)=\bigtriangleup(\bm{x},\bm{c_i}) - r_i.
\end{equation}
\end{mydef}

According to Definition \ref{mydef3}, the distance between any queried sample and the GB is essentially the distance from the sample to the surface of the ball.

Therefore, as with the traditional $k$NN, for GB$k$NN, the label of each queried sample is determined by $k$ surrounding training samples. The difference is that the parameter $k$ is adaptively determined by the number of samples contained in the nearest GB, which is particularly important for the application and promotion of $k$NN.

\section{GBG++ Method}
\label{sec:GBG++}
\noindent As discussed, the existing GBG methods rely on $k$-means or $k$-division, which both require random centers and need to calculate the distances between each center and all samples. In other words, the existing GBG methods are unstable and time-consuming. Moreover, existing GBG methods consider it reasonable that a single sample generates a GB; however, most of these samples are far from their peers, thus being outliers.

\begin{figure*}[htbp]
\vspace{-0.3in}
\centering
\includegraphics[height=3.8in,width=7in]{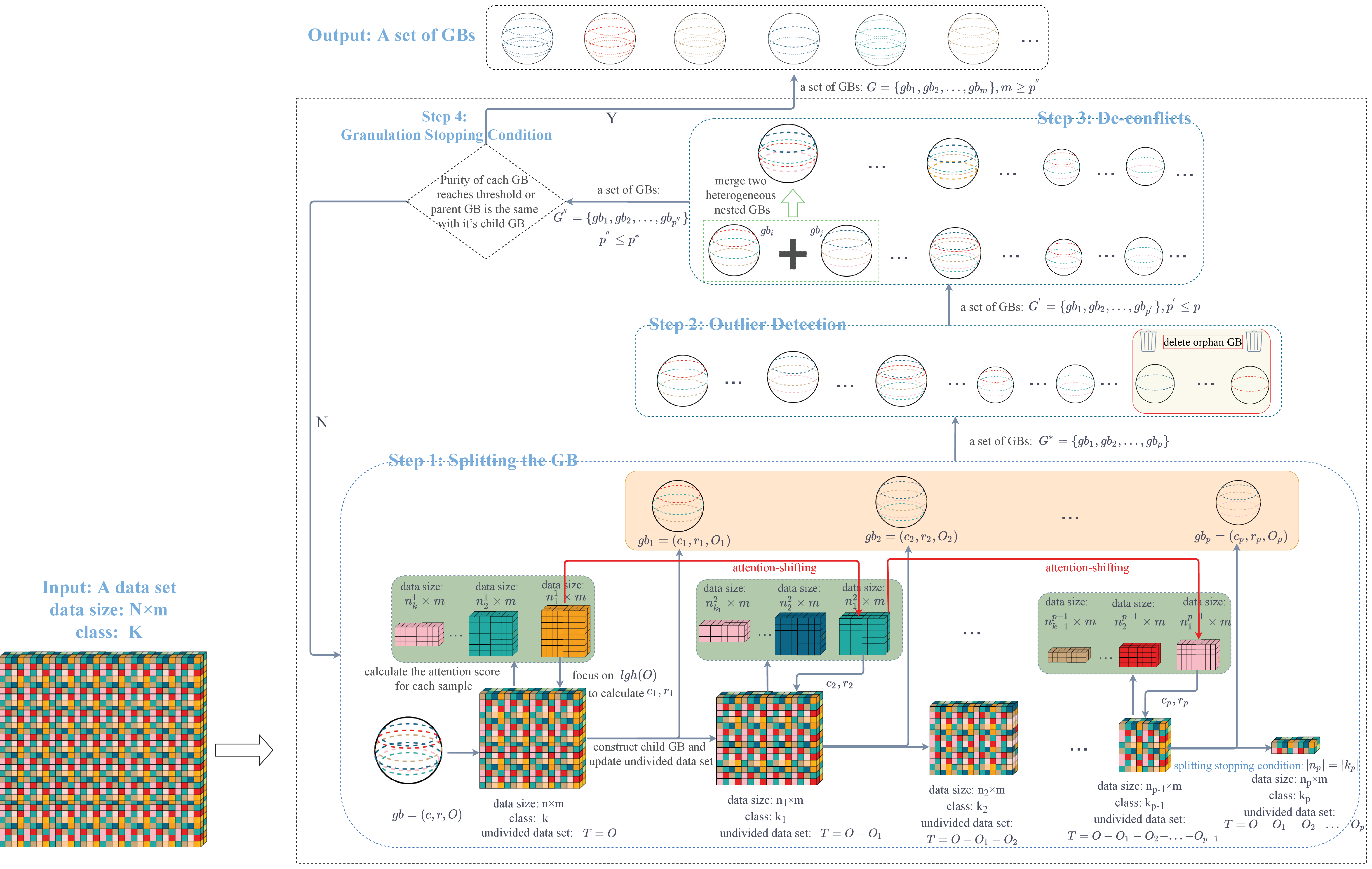}
\caption {Architecture of GBG++ method.}
\label{fig4}
\vspace{-0.1in}
\end{figure*}

Consequently, GBG++ is proposed to address the aforementioned problems. In detail, considering that the AM is data-driven, effective, and efficient, a method for splitting the GB based on the AM is proposed. Moreover, outlier detection and de-conflicts are performed for the set of GBs obtained by the splitting method. Subsequently, the granulation stopping condition is applied for the entire granulation process. Finally, the detailed design of the GBG++ method is presented. Fig. \ref{fig4} shows the architecture of the proposed GBG++ method.

\subsection{Splitting the GB Based on AM}
\label{subsec:split}
The core idea of the proposed method for splitting the GB is to iteratively construct finer GBs on the majority class of undivided samples. The center and radius of finer GB are calculated on the majority class. Step 1 of Fig. \ref{fig4} shows that attention is on undivided samples and the majority class. As GBs are constructed in order, attention shifts. Several definitions are given below to describe the GB splitting process.

As mentioned previously, when the purity threshold is given, the sample size also influences the quality of the GB. Thus, the Definition \ref{mydef1} is redefined as below.

\begin{mydef}\label{mydef4} Given a dataset $D$. Suppose $G$ be a set of GBs generated on $D$, in which $gb_i=(\bm{c_i},r_i,O_i)$ is generated on $D_i(D_i\subseteq D, D_i \neq \emptyset)$. For $\forall gb_i \in G$, the set of samples contained is defined as follows,
\begin{equation}\label{eq9}
O_i = \{(\bm{x},y)\in D_i|\bigtriangleup(\bm{x},\bm{c_i})\leq r_i\},
\end{equation}
\end{mydef}
where the definitions of $\bm{c_i}$ and $r_i$ refer to Definition \ref{mydef1} and $i=1,2,...,m$.

\begin{mydef}\label{mydef5} Given a GB $gb=(\bm{c},r,O)$. If $\exists G^*=\{gb_i|gb_i=(\bm{c_i},r_i,O_i), \cup_{i=1}^l O_i \subseteq O, \cap_{i=1}^l O_i = \emptyset,i=1,2,...,l\}$ and $l<m$, then $G^*$ is the set of child GBs of $gb$ and $\forall gb_i(gb_i \in G^*)$ is called the child GB of its parent GB $gb$.
\end{mydef}

According to Definition \ref{mydef5}, the parent GB represents the split GB, and the child GB represents the GB obtained by splitting its parent. Obviously, the samples contained in the child GB are inherited from its parent. In particular, in the process of GBG, if $l=1$ and $O = O_i$, then parent GB $gb$ reaches a stable state and does not need to be split.

In the field of GrC, the granulation strategy should ensure that the constructed IGs can effectively describe the distribution characteristics of the original dataset. Meanwhile, for GBC, the greater the purity of the GB, the better the description of the original dataset by GBs. In addition, with a certain purity, the more samples contained in the GB, the higher its quality. In general, the data distribution follows certain rules; for example, homogeneous samples are closely distributed. Thus, when splitting the GB, if the majority class of undivided samples is processed first, then the constructed child GB can contain more samples of higher purity. Besides, redundant calculations can be avoided if only one homogeneous sample type is processed each time.

The core idea of hard attention in the AM is to select the element that is most relevant to a query element in a given input and distribute attention to that element. The working principle of hard attention is briefly described below.

Given a GB $gb=(\bm{c},r,O)$, the $T=\{(\bm{x_1},y_1),(\bm{x_2},y_2),...,$ $(\bm{x}_{|T|},y_{|T|})\}(T \subseteq O)$ is the set of undivided samples, and $\bm{Y}^*=[y_1,y_2,...,y_{|T|}]$ is the corresponding label vector. Then, the attention score $\alpha_i$ of each sample in $T$ can be expressed by the formula
\begin{equation}\label{eq14}
\alpha_i=\begin{cases}
    1, & \text{if } i= {\underset{j}{{\arg\max} \,} (S(y^{'},y_j))}, \\
    0, & \text{otherwise},
\end{cases}
\end{equation}
where $y_j\in \bm{Y}^*$, $S(y^{'},y_j)$ represents the similarity between $y_j$ and $y^{'}$ which is the mode of the elements in $\bm{Y}^*$, and $j=1,2,...,|T|$.

Therefore, when splitting the GB, priority is given to samples with an attention score of 1, and the center and radius of the child GB are determined on these samples based on Definition \ref{mydef4}. To facilitate the introduction of the process of splitting the GB, samples with an attention score of 1 are represented by the majority class, which is defined as follows,
\begin{equation}\label{eq10}
lhg(T)=\{(\bm{x_i},y_i)|\alpha_i = 1\}.
\end{equation}
Thus, from the perspective of AM, the majority class of the undivided dataset $T$ is given priority in attention, and the samples in the complementary set of $lhg(T)$ are temporarily ignored; namely, their weights are set to 0.

As shown in Step 1 of Fig. \ref{fig4}, for any $gb=(\bm{c},r,O)$ that need to be split, the steps of constructing any child GB by the proposed splitting method can be briefly summarized as follows. First, the undivided dataset $T$ is initialized as $O$, namely, $T=O$. Second, based on Eq.\ref{eq14} and Eq.\ref{eq10}, the attention is on majority class $lhg(T)$. Third, based on Definition \ref{mydef4}, the $\bm{c_1},r_1$ of child GB $gb_1=(\bm{c_1},r_1,O_1)$ is constructed on $lhg(T)$ and $O_1$ is obtained based on $O$, and the undivided dataset $T$ is updated to $O-O_1$. Subsequently, the attention is shifted to the majority class $lhg(O-O_1)$ to generate another GB iteratively until the splitting stopping condition is satisfied. Finally, the child GB set $G^* = \{gb_1,gb_2,...,gb_p\}$ is generated.

Obviously, owing to the strategy of `priority for majority class', the constructed child GBs can be with higher purity while containing more samples. Thus, compared to existing GBG methods, the entire granulation iteration process of the proposed GBG++ can be stopped earlier. Besides, some redundant distance calculations are avoided to improve the efficiency of the GBG method.

\vspace{-0.3cm}

\subsection{Outlier Detection and Splitting Stopping Condition}
\label{subsec:ODSSC}
Generally, the samples within the same GB are closely distributed. Therefore, if a GB contains only one sample, the sample can be considered an outlier. Furthermore, according to the splitting method introduced in Section \ref{subsec:split}, for any GB, its child GBs are iteratively constructed. A reasonable iteration stopping condition not only improves the quality of the child GBs but also improves the efficiency of the splitting method. If the undivided samples are outliers, the splitting iterations are stopped directly. An outlier detection method is introduced first.

Let $gb=(\bm{c},r,O)$ be the GB. If $|O|=1$, then $gb$ is called an orphan GB. During the process of splitting the GB, the reason for producing orphan GBs can be briefly described as follows. Each GB(except for the initial GB) is constructed on the majority class of the undivided samples of its parent GB. Specifically, the construction method is to construct a ball with the mean feature vector of all samples in the majority class as the center $\bm{c}$ and the mean value of the distances from the samples in the majority class to $\bm{c}$ as the radius. If the distances from all the homogeneous samples to $\bm{c}$ are regarded as a variable $\bm{d}$, and its variance $var(\bm{d})$ can be calculated. According to the properties of mean and variance, when $var(\bm{d})=0$, all samples are divided on the surface of the GB. When $var(\bm{d})>0$, some samples are divided into the ball or on the surface, and some samples must fall outside the GB. Therefore, when a sample is quite different from the others in this majority class, it falls outside the ball and eventually be classified into other GBs or forms the orphan GB in subsequent granulation iterations. It indicates that the sample is significantly alienated from the majority of samples and that no companions can be found in subsequent granulation iterations. Therefore, if an isolated sample occurs, it can be regarded as an outlier. In other words, the orphan GB can be regarded as an outlier. As in Step 2 of Fig. \ref{fig4}, for the child GB set $G^* = \{gb_1,gb_2,...,gb_p\}$ obtained in Step 1, check the number of samples contained in each $gb_i,i=1,2,...,p$, and drop the GB containing only one sample, that is, the orphan GB which can be regarded as the outlier.

Given a GB $gb=(\bm{c},r,O)$, the $T(T \subseteq O)$ is the set of undivided samples, and the $lhg(T)$ is majority class of $T$. And, $gb_1=(\bm{c_1},r_1,O_1)$ is generated on $T$. Then, local homogeneous outliers can be defined as follows,
\begin{equation}\label{eq11}
lho(T) = \{(\bm{x},y)\in lhg(T)|\bigtriangleup(\bm{x},\bm{c_1}) > r_1\}.
\end{equation}

According to Eq.\ref{eq11} and Definition \ref{mydef4}, for a set of homogenous samples, a local outlier is a sample that is outside the $gb_1$. In other words, local homogenous outliers deviate from the majority of samples. If multiple homogenous outliers exist, they may be divided into new GBs in subsequent granulation iterations. Therefore, the qualifier `local' is used here. However, when the undivided samples are heterogeneous, the GBs constructed in subsequent splitting iterations are all orphan GBs; thus, subsequent splitting iterations are not necessary.

In summary, the core idea of the proposed outlier detection method is to identify orphan GBs. The splitting stopping condition is that the undivided samples can only be constructed as orphan GBs; that is, as shown in Step 1 of Fig. \ref{fig4}, the splitting stopping condition can be expressed as the number of undivided samples is equal to the number of label category.

\vspace{-0.3cm}

\subsection{De-Conflicts}
\label{subsec:De-Conflicts}
Since the proposed GBG method constructs GBs in order, the constructed GBs may be nested with each other.

Given two GBs $gb_1=(\bm{c_1},r_1,O_1)$ and $gb_2=(\bm{c_2},r_2,O_2)$. The $gb_1$ and $gb_2$ are called the nested GBs, if
$$\bigtriangleup(\bm{c_1,c_2}) \leq abs(r_1-r_2),$$ where $abs(\ast)$ denotes the absolute value of $\ast$.

Suppose $l_1$ and $l_2$ be the labels of $gb_1$ and $gb_2$ respectively. If $l_1 \neq l_2$, then $gb_1$ and $gb_2$ are called heterogeneous nested GBs, otherwise called homogeneous nested GBs.

\begin{figure}[htbp]
\centering
        \begin{minipage}[t]{0.45\linewidth}
        \centering
        \includegraphics[height=3.0cm, width=3.8cm]{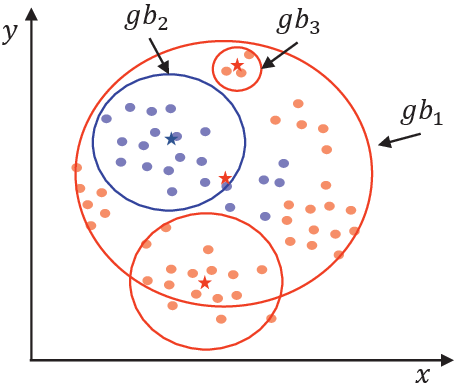}
        \small (a) Conflicts between GBs.
        \end{minipage}%
        \begin{minipage}[t]{0.45\linewidth}
        \centering
        \includegraphics[height=3.0cm, width=3.8cm]{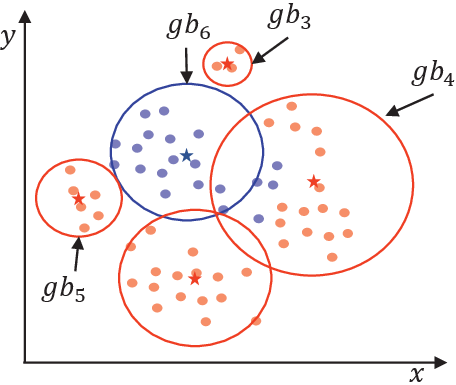}
        \small (b) Performing De-conflicts.
        \end{minipage}%

\caption {Example of conflicts and de-conflicts.}
\label{fig6}
\vspace{-0.3cm}
\end{figure}

For example, as illustrated in Fig.\ref{fig6}(a), as the distribution of samples contained in $gb_1$ is relatively dispersed, the $gb_1$ is nested with two previously constructed GBs, namely, $gb_2$ and $gb_3$. The labels of $gb_1$ and $gb_3$ are the same, called homogeneous nested GBs. Meanwhile, as the labels of $gb_1$ and $gb_2$ are different, they are called the heterogeneous nested GBs, that is, the GBs conflict.

Mutual exclusion exists between two GBs with the heterogeneous nested relation, indicating that at least one GB is inconsistent with the data distribution of the original dataset. Generally, heterogeneous nested GBs are distributed near the class boundary of the original dataset. This implies that GBs may fail to accurately depict the class boundary. Besides, if there is no nested relationship between any two GBs, then there will be no nested relationship between their child GBs.
Therefore, in each granulation iteration, if the operation to remove the nested relationship is applied, the nested relationship can be eliminated as soon as possible while the efficiency of the GBG method is not significantly affected.

Therefore, the heterogeneous nested GBs are merged to be split again at each granulation iteration in the proposed GBG++ method, as shown in Step 3 of Fig. \ref{fig4}. In Step 3, for the child GB set $G^{'} = \{gb_1,gb_2,...,gb_{p^{'}}\}$ obtained after the outlier detection in Step 2, check whether there is a conflict relationship between the child GBs. If any two GBs $gb_i$ and $gb_j$ conflict, merge them to generate a new GB using Definition \ref{mydef4} and drop $gb_i$ and $gb_j$ at the same time, where $i,j=1,2,...,p^{'}$ and $i\neq j$. In addition, as in Fig.\ref{fig6}(a), the $gb_1=(\bm{c_1},r_1,O_1)$ and $gb_2=(\bm{c_2},r_2,O_2)$ are with heterogeneous nested relationship, which are merged to be split again. In detail, first, merge the samples contained in $gb_1$ and $gb_2$ to obtain $O_1 \cup O_2$. Second, based on Definition \ref{mydef4}, construct a new GB on the $O_1 \cup O_2$ and calculate its label and purity based on Definition \ref{mydef2}. Third, the new GB will go to Step 4, as in Fig.\ref{fig4}, if the purity of the new GB does not reach the threshold, it will return to Step 1 of Fig. \ref{fig4} to continue being split. In addition, both $gb_1$ and $gb_2$ are deleted simultaneously. Finally, as shown in Fig.\ref{fig6}(b), the new GBs, $gb_4$, $gb_5$ and $gb_6$, are no longer in conflict.

Generally, if two homogeneous GBs are nested, it means that the distribution of these homogeneous samples is not so tight. Although homogenous nested GBs do not affect the data distribution characteristics of the original dataset, they can reflect its discrete characteristics. Therefore, for the sake of high efficiency, homogeneous nested GBs are not processed.

\subsection{Granulation stopping Condition and Algorithm Design}
\label{subsec:GSCAD}
Based on the method of splitting GB introduced in Section \ref{subsec:split}, since the heterogeneous nested GBs are merged and split again, as shown in Section \ref{subsec:De-Conflicts}, then a few GBs will be stable, even if they do not reach the threshold. Thus, as shown in Step 4 of Fig. \ref{fig4}, the GBG++ method still needs to determine whether GB can stop granulation by judging whether the purity of GB reaches the threshold and also needs to judge whether GB reaches a stable state, that is, whether the GB is the same as its parent. Especially for the second part of the condition, since the GB can no longer reach the threshold, it must be removed directly.
For the entire execution process of GBG++, starting from taking the entire training dataset as the initial GB, iterate steps 1-4 until all the constructed GBs meet the judgment conditions shown in Step 4, then the granulation stops, and the final target GB set $G=\{gb_1,gb_2,...,gb_m\}$ is obtained.

\begin{figure}[htbp]
\vspace{-0.3cm}
	\centering
	\begin{minipage}[t]{0.5\linewidth}
		\centering
		\includegraphics[height=3.3cm, width=4cm]{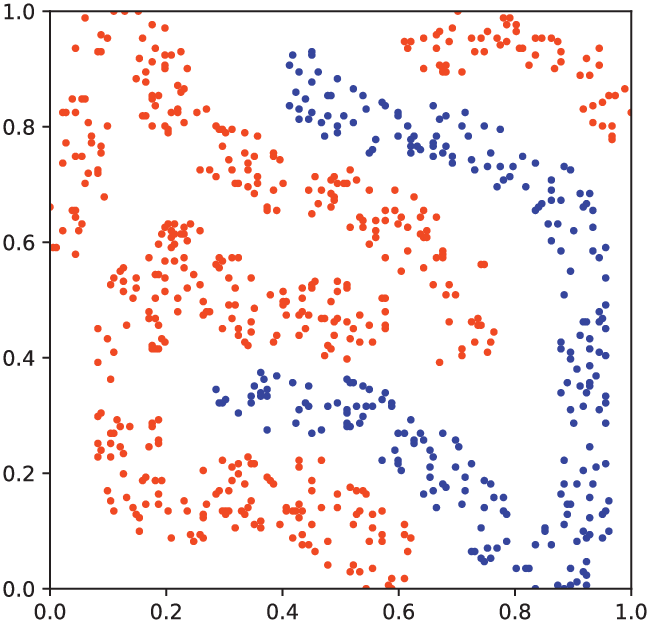}
		\small (a) Original dataset.
	\end{minipage}%
	\begin{minipage}[t]{0.5\linewidth}
		\centering
		\includegraphics[height=3.3cm, width=4cm]{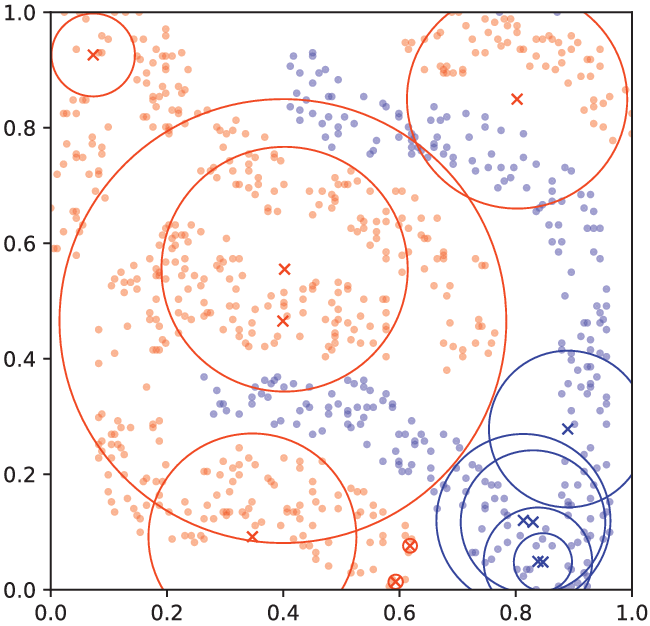}
		\small (b) $1th$ iteration of granulation process.
	\end{minipage}%
    \quad
	\begin{minipage}[t]{0.5\linewidth}
		\centering
		\includegraphics[height=3.3cm, width=4cm]{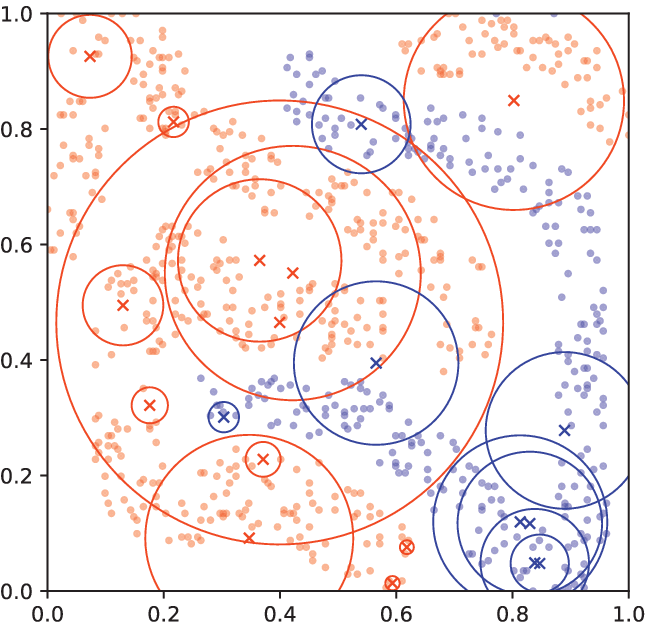}
		\small (c) $2th$ iteration of granulation process.
	\end{minipage}%
	\begin{minipage}[t]{0.5\linewidth}
		\centering
		\includegraphics[height=3.3cm, width=4cm]{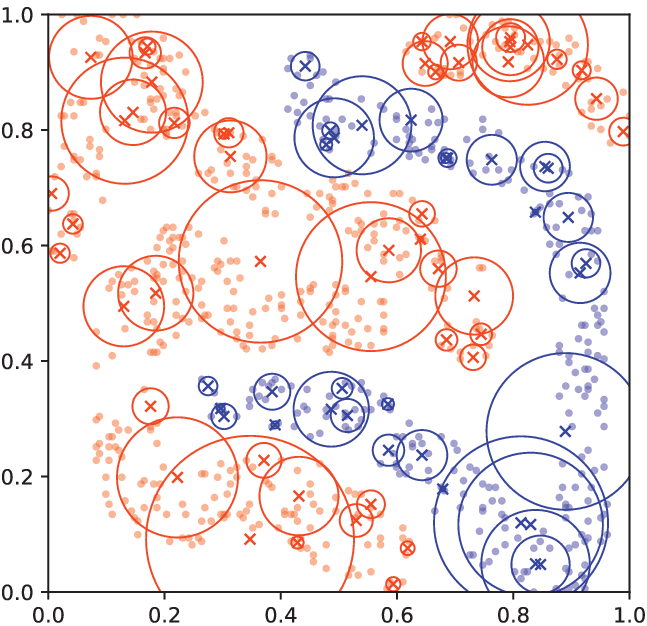}
		\small (d) Last iteration of granulation process.
	\end{minipage}%
    \quad
    \begin{minipage}[t]{0.5\linewidth}
	\centering
	\includegraphics[height=3.3cm, width=4cm]{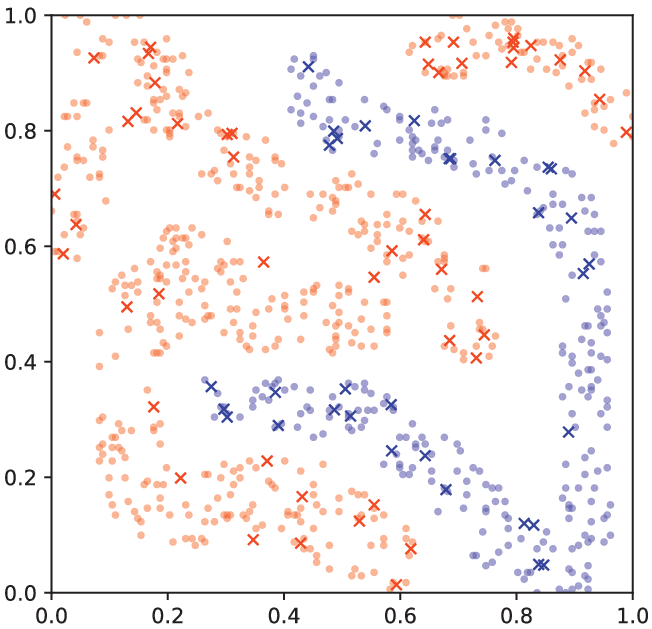}
	\small (e) All GBs only with centers.
    \end{minipage}%
    \begin{minipage}[t]{0.5\linewidth}
	\centering
	\includegraphics[height=3.3cm, width=4cm]{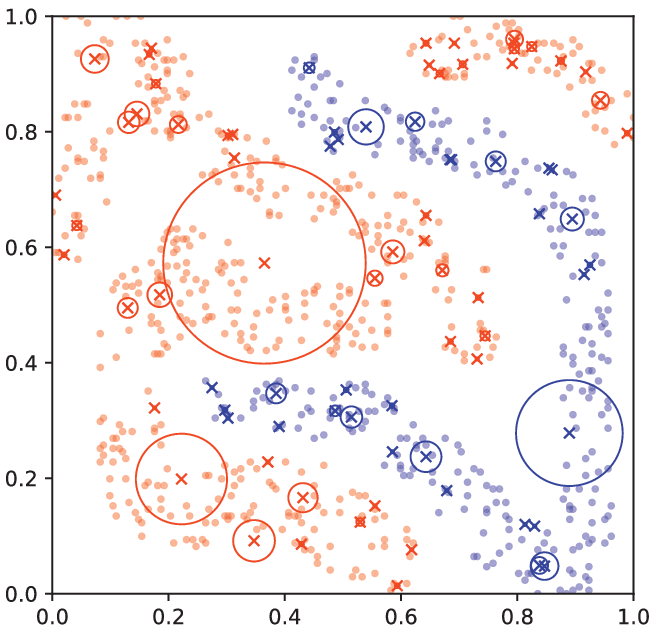}
	\small (f) All reconstructed GBs.
    \end{minipage}%
\caption {Process of generating GBs by GBG++ method.}
\label{fig7}
\vspace{-0.1cm}
\end{figure}

An example is provided below to visualize the entire execution process of the GBG++ method. In detail, the Fig. \ref{fig7}(a) shows the whole original dataset fourclass \cite{45}. Fig. \ref{fig7}(b) and (c) show all the generated GBs in the $1th$ and $2th$ granulation iterations, respectively. Fig. \ref{fig7}(d) shows all the constructed GBs when all GBs meet the granulation stopping condition. Fig. \ref{fig7}(d) indeed shows that taking the geometric information of GBs to characterize the data distribution characteristics of the original dataset. Moreover, notably, a larger radius of GB does not indicate that it contains more samples. As shown in Fig. \ref{fig7}(e), only the centers of GBs are used to determine the data distribution characteristics, which can effectively depict the complexity of class boundaries for data distribution. In essence, if other conditions are consistent, the more samples contained in the GB, the larger its radius. For classification tasks, the more samples that support the classification rules, the higher the quality of the classification rules. Thus, the proportion of samples contained in GB to all samples might be used as a measure of GB's size or quality when other conditions remain unchanged in GB-based classification tasks. The GB reflecting the sample size is used to describe the distribution of the original dataset, and the schematic diagram is shown in Fig. \ref{fig7}(f). It can be seen that Fig. \ref{fig7}(f) describes the original dataset more accurately than Fig. \ref{fig7}(d) and (e).
Moreover, the design of the proposed GBG++ method is presented in Algorithm \ref{alg1}.

\begin{algorithm}[htbp]
\caption{GBG++ Method.}
\label{alg1}
\SetKwInOut{Input}{Input}\SetKwInOut{Output}{Output}
\Input {Dataset $D$, purity threshold $P$.}
\Output {A set of GBs $G$.}
Treat the entire $D$ as the initial GB $gb_i^j$($gb_i^j=(\bm{c_i^j},r_i^j,O_i^j)$), where $i$ represents the number of iterations initialized to 1, and $j$ represents $ith$ GB generated in each iteration initialized to 1\;
Initialize $G\leftarrow \emptyset$\;
\Repeat{Cardinality of GB set $|G|$ does not increase}{
    \For {each GB $gb_i^j\in G$}{
        Calculate purity $p_i^j$ by Eq. \ref{eq4}\;
        \If {$p_i^j < P$}{
            $T$ represents the set of undivided samples in $gb_i^j$, and $gb_{i+1}^k$ denotes $kth$ child GB in child GB set $G^*$\;
            Initialize $T \leftarrow O_i^j$, $G^*\leftarrow \emptyset$,  and $k\leftarrow\ 1$\;
            Calculate $lgh(T)$ by Eq.\ref{eq10}\;
            \While {$|lhg(T)|!=1$}{
                Construct $gb_{i+1}^k$ on $T$ by Definition \ref{mydef4}\;
                $T\leftarrow T-O_{i+1}^k$, $k\leftarrow k+1$\;
                $G^*\leftarrow G^* +\{gb_{i+1}^k\}$\;
                Calculate $lhg(T)$\;
            }
            \If {$k==1$ and $|O_i^j|==|O_{i+1}^k|$}{
                $G\leftarrow G-\{gb_i^j\}$\;
                \textbf{continue}\;
            }
            \If {$k>1$}{
                Perform de-conflict on $G^*$\;
            }
            $G\leftarrow G+G^*$\;
            $G\leftarrow G-\{gb_i^j\}$\;
        }
    }
}
Return $G$.
\end{algorithm}

\subsection{Time Complexity}
\label{subsec:Time}
The time complexity of original GBG method\cite{29} and the acceleration GBG method\cite{32} is $O(ktmN)$ and $O((tk-t+gs+1)N)$ respectively, where $m$ represents the number of iterations in $k$-means, $t$ represents the number of iterations of the GBG method, $k$ indicates the number of label categories and $g$ represents the number of GBs. The convergence speed of the existing GBG method is fast and can be considered approximately linear.

Suppose a dataset that contains $N$ samples and $k$ classes. Assume that the entire granulation process is iterated $t$ rounds. Let $r_i(i=1,2,...,s)$ represent the number of samples that are repeatedly calculated in the $ith$ granulation iteration; that is, the sample is used to calculate the distances to multiple centers. In the first round, the initial GB is split, and the time complexity is $O(N+r_1)$. In the second round, the GBs constructed in the previous round are split, and the time complexity is $O(N+r_2)$. For the third round, the time complexity is $O(N+r_3)$. Suppose that the total number of constructed GBs is $m$ after $t$ iterations. Subsequently, the center and radius of the $m$ GBs are calibrated, and the time complexity is $O(N)$. Notably, because the time complexity of performing the de-conflict is constant after each iteration, it can be ignored. In conclusion, the total time complexity of the proposed GBG++ method is $O((s+1) N+(r_1+r_2+...+r_s))$. Owing to that, in each round, $r_i\ll N$, the time complexity can be further approximated as $O((t+1)N)$. In addition, most of the GBs reach the granulation stopping condition in the middle of iteration, so the actual time complexity is far less than $O((t+1)N)$. Therefore, the time complexity of the GBG++ method can be considered approximately linear, which is lower than that of the existing GBG method.

\section{GB$k$NN++}
\label{sec:GB$k$NN++}
\noindent The core idea of the existing GB$k$NN is to calculate the distances from a queried sample to all GBs, and the queried sample shall be labeled with the nearest GB's label. However, similar to the common problem of existing GB-based classifiers, the GB$k$NN only considers the geometric characteristics of the GB but ignores the number of samples contained in the GB when constructing the classification rule.

\begin{figure}[htbp]
\vspace{-0.3cm}
\centering
\includegraphics[height=1.5in,width=2.0in]{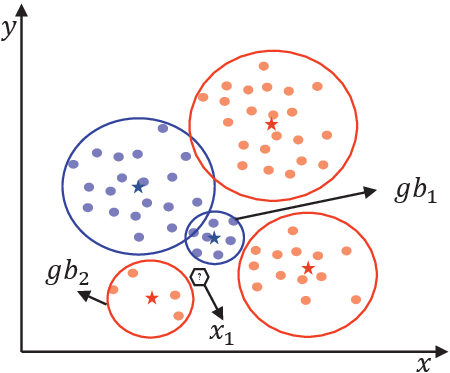}
\caption {Misclassification at class boundary.}
\label{fig8}
\vspace{-0.3cm}
\end{figure}

According to Definition \ref{mydef3}, the existing GB$k$NN is sensitive to the GB's radius. As previously mentioned, a larger GB's radius does not indicate higher quality. For instance, as shown in Fig.\ref{fig8}, for the queried sample $\bm{x_1}$ whose true label is $1$ distributed at the class boundary. According to the original GB$k$NN, the distance from $\bm{x_1}$ to $gb_2$ whose label is $0$ is minimum, then$\bm{x_1}$ is labeled as 0. Obviously, the queried sample is incorrectly classified. Compared with $gb_1$, the density of sample distribution in $gb_2$ is lower, while its radius is larger. Thus, if the $gb_1$ and $gb_2$ are differentiated in terms of sample size, the misclassification can be optimized. In addition, the more samples contained in the GB, the higher the support degree of the classification rule constructed by the GB.

Therefore, an improved GB$k$NN is proposed, which is denoted as GB$k$NN++. The classification rule of the GB$k$NN++ for the queried sample is to calculate the harmonic distance from the queried sample to each GB and label the queried sample with the corresponding label of the GB with the minimum harmonic distance. The definition of harmonic distance is defined as follows,

\begin{mydef}\label{mydef7} Given a dataset $D$. Suppose $G$ be a set of GBs generated on $D$. For any queried sample $\bm{x}$, the harmonic distance between $\bm{x}$ and $\forall gb_i (gb_i=(\bm{c_i},r_i,O_i),gb_i \in G,i=1,2,...,m)$ is defined as below,
\begin{equation}\label{eq12}
hd(\bm{x},gb_i)= \bigtriangleup(\bm{x},\bm{c_i}) - \frac{|O_i|}{\sum_{j=1}^{m}|O_j|}.
\end{equation}
\end{mydef}

According to Definition \ref{mydef7}, the harmonic distance consists of two parts. The first part refers to the distance from the queried sample to the center of the GB, which considers the geometric information of GB. The second part is a penalty term which represents the quality of GB, that is, the proportion of samples contained in $gb_i$ to samples contained in all GBs. Based on the harmonic distance, if the distances between the queried sample and the centers of two GBs are the same, the sample will be labeled as the GB that contains more samples.
In addition, the time complexity of GB$k$NN++ is negligible relative to that of the GBG++ method since the input space dimensions are significantly reduced. Additionally, the time complexity of the GB$k$NN++ is $O(g)$, which is considerably smaller than $k$NN, where $g$ is the number of GBs. Moreover, the algorithm design of the GB$k$NN++ is presented in Algorithm \ref{alg2}.

\begin{algorithm}[!ht]
\caption{GB$k$NN++.}
\label{alg2}
\SetKwInOut{Input}{Input}\SetKwInOut{Output}{Output}
\Input {A GB set $G$ generated by Algorithm \ref{alg1} and the queried sample $\bm{x}$.}
\Output {Label $l$ of $\bm{x}$.}
    Calculate the harmonic distance between sample $\bm{x}$ and each $gb_i(i=1,2,...,m)$ by Eq. (\ref{eq12})\;
    Label sample $\bm{x}$ with the nearest GB's label $l$\;
Return $l$.
\end{algorithm}
\vspace{-0.5cm}

\section{Experimental Results and Analysis}
\label{sec:Experiments}
\noindent In this section, the experimental settings are introduced first, including verification procedures and evaluation metrics, used datasets, comparison methods and hyperparameter settings. Then, a detailed introduction to the performance of the proposed method in terms of effectiveness, efficiency, stability, and robustness is provided. Finally, the parameter sensitivity analysis is given for the hyperparameters involved in the proposed method. The ablation study is provided to verify the role of each module of the proposed method. What's more, the hardware environment information is 3.00GHz Intel i9-10980XE CPU, and the software environment information is Python 3.9.7.

\subsection{Experimental Settings}
\label{subsec:ExperSettings}

\subsubsection{Validation Procedure and Evaluation Metrics}
\label{subsubsec:ValiProce}

\noindent To comprehensively demonstrate the proposed method, considering that the proposed GBG++ can be regarded as a data preprocessing method and the proposed GB$k$NN++ is essentially a classifier, the experiments are designed as a multi-class classification task. The proposed method is validated by the accuracy of classification results, the time cost, the anti-noise ability, and the stability. In order to alleviate the false performance gain of the model that may be caused by overfitting, the 10-fold cross-validation method is employed in each subsection in Section \ref{sec:Experiments} to measure the actual generalization ability of the proposed method.

The $Accuracy$ is taken as the basic evaluation metric, which is classic and commonly used in supervised learning. Test $Accuracy$ refers to the ratio of the number of correctly predicted test samples to the number of total test samples. Notably, in this paper, test $Accuracy$ represents the average result of 10-fold cross-validation. In addition, the standard deviation ($Sd$) of test $Accuracy$ of each fold is used to measure the consistency of the classifier's performance on the same dataset when 10-fold cross-validation is employed.

Furthermore, to further verify that there are significant differences between classifiers, the Wilcoxon rank-sum test\cite{51} is employed. The following hypotheses are tested in this Section. The null hypothesis is that the proposed classifier is not significantly different from the comparison methods, while the alternative hypothesis is the opposite.

Let $d_i,i=1,2,...,N$ denote the differences between the test $Accuracy$ of GB$k$NN++ and comparison methods on the $ith$ dataset. The differences are ranked based on their absolute values. If there are any ties in the rankings, average ranks are assigned. Let $R^+$ denote the sum of ranks of comparison methods that outperform GB$k$NN++ and $R^-$ denote the sum of ranks of GB$k$NN++ outperforming comparison methods.The definitions of $R^+$ and $R^-$ are as follows,

\begin{equation}\label{eq7}
R^+ = \sum_{d_i>0} rank(d_i) +\frac{1}{2}\sum_{d_i=0} rank(d_i),
\end{equation}

\begin{equation}\label{eq8}
R^- = \sum_{d_i<0} rank(d_i) +\frac{1}{2}\sum_{d_i=0} rank(d_i).
\end{equation}
Let $T(T=min(R^+,R^-))$ denote the smaller of $R^+$ and $R^-$.

Besides, considering the possibility of visualization, to verify the advantages of the proposed method's efficiency, the logarithm of the average training time $t$ (unit: ms) of the classifier is used to evaluate the efficiency when 10-fold cross-validation is employed, which is as follow.

\begin{equation}\label{eq13}
LNT = \ln(t).
\end{equation}

\subsubsection{Datasets}
\label{subsubsec:Datasets}

\noindent Twenty benchmark datasets whose detailed information is shown in Table \ref{table1} are used to evaluate the proposed methods. These datasets come from various sources, such as the UCI Repository of Machine Learning datasets\cite{38}, scikit-feature feature selection repository\cite{57}, and image classification datasets\cite{41,42,43}. These datasets include all different types of datasets, such as high-dimensional datasets, low-dimensional datasets, binary datasets, multi-class datasets, large-scale datasets, and small-scale datasets. In terms of application scenarios, these datasets involve text data, image data, biological data, medical data, and audio data.
Notably, unstructured datasets involved have been transformed into structured numerical datasets by providers. In order to alleviate the impact of dimensions on the classifier's learning process, all used datasets are normalized.

\begin{table}[htbp]
	\renewcommand{\arraystretch}{1.0}
	\caption{The Details of Datasets.}
	\label{table1}
	\centering
	\renewcommand\tabcolsep{3.6pt}
	\begin{tabular}{lcccc}
		\toprule[1pt]
		{Datasets} &{Samples} & {Features}& {Classes} & {Sources}
        \\ \hline
        CLL\_SUB\_111           &111    &11340  &3       & \cite{57}    \\
		Parkinsons            &195	  &22	  &2       & \cite{38}     \\
        lung                  &203    &3312   &5       & \cite{57}    \\
		Sonar                 &208	  &60	  &2       &  \cite{38}    \\
		Ecoli                 &336	  &7	  &8       &  \cite{38}    \\
		ORL             &400	  &1024	&40      & \cite{40}     \\
		Credit Approval(Credit)        &690	  &15	  &2       &  \cite{38}      \\
		Diabetes              &768	  &8	  &2       &  \cite{38}      \\
		BreastMNIST           &780	  &784	&2       &   \cite{41}     \\
		fourclass             &862	  &2	  &2       &  \cite{45}      \\
		splice                &979	  &60   &2       &   \cite{38}     \\
        RELATHE               &1427   &4322   &2       & \cite{57}    \\
		COIL20                &1440	  &1024	&20      &  \cite{39}    \\
		Isolet                &1560	  &617	&26      &  \cite{57}    \\
		Image Segmentation(Image-Seg)    &2310	  &19	  &7    &  \cite{38}        \\
		Page Blocks(Page)           &5473	  &10	  &5    &  \cite{38}        \\
		svmguide1             &7089	  &4	  &2    &    \cite{46}      \\
		COIL100               &7200	  &1024	&100  &     \cite{39}     \\
		Pen                   &10992	&16	  &10     & \cite{38}       \\
		OrganMNIST-Sagittal(OrganMNIST)   &25221	&784	&11     &   \cite{41}     \\
		Cod-RNA                &49466	&8	  &2      &   \cite{44}     \\
		MNIST                 &70000	&784	&10     &   \cite{42}     \\
		Fashion-MNIST         &70000	&754	&10     &  \cite{43}      \\
		Skin Segmentation(Skin-Seg)     &245057	&3	  &2    &  \cite{38}        \\
		\toprule[1pt]
	\end{tabular}
\vspace{-0.5cm}
\end{table}

\subsubsection{Comparison Methods and Hyperparameter Settings}
\label{subsubsec:CompMethHyperparaSetting}

\noindent The proposed method is compared with $2$ state-of-the-art GB-based classifiers and $3$ classic machine learning classifiers whose hyperparameter settings are as follows.

\begin{itemize}
\item The original GB$k$NN(ORI-GB$k$NN)\cite{29}: the setting of $k$ of $k$-means used in ORI-GB$k$NN is as introduced in Section \ref{subsec:Ori-GBG}. And, the setting of $k$ of $k$-NN used in ORI-GB$k$NN is as introduced in Section \ref{subsec:Ori-GB$k$NN}.

\item An acceleration GB$k$NN(ACC-GB$k$NN)\cite{32}: the setting of $k$ of $k$-division used in ACC-GB$k$NN is as introduced in Section \ref{subsec:ACC-GBG}.

\item $k$-Nearest Neighbor($k$NN): the $k$ is set to $1, 3, 5, 7, 9, 11,$ $ 13$ and $15$ respectively, and the average level of classification results are as the final result.

\item Support vector machine(SVM): the kernel function uses the radial basis function, and its hyperparameter is set to the reciprocal of the number of features of the dataset. The other hyperparameters are consistent with the default parameters in scikit-learn, which is a popular open-source machine-learning library for Python.

\item Classification and regression tree(CART): all hyperparameters of CART are consistent with the default parameters in scikit-learn.
\end{itemize}

Notably, with increasing purity, taking GBs to describe the original dataset is generally more accurate. Therefore, the purity threshold might be set to $1.0$ without loss of generality for all GB-based classifiers including the proposed GB$k$NN++.

\begin{figure*}[htbp]
\vspace{-0.5cm}
\centering
        \begin{minipage}[t]{0.45\linewidth}
        \centering
        \includegraphics[height=1.5in,width=2.5in]{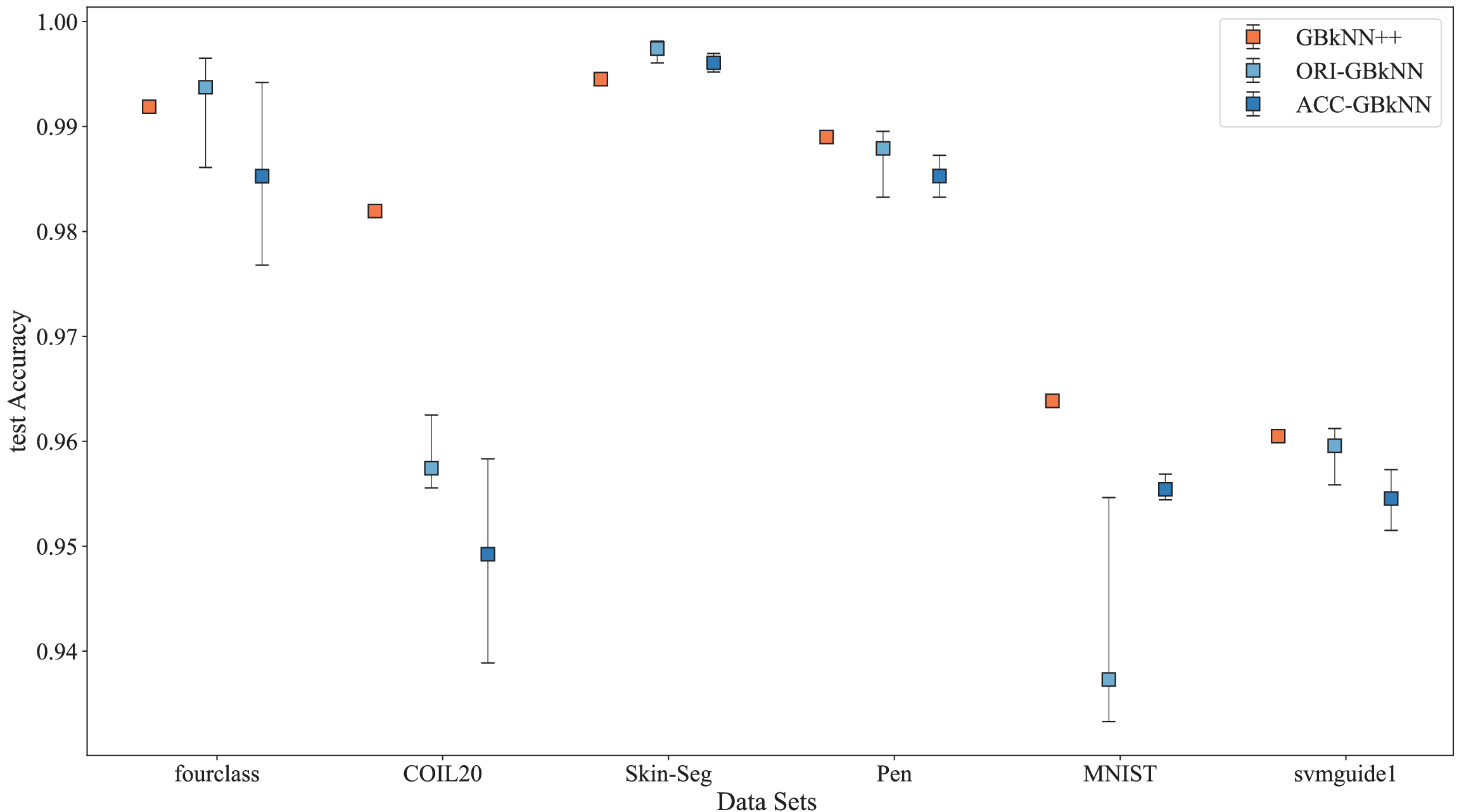}

        \small{(a) Stability on test $Accuracy$ on $1th$ group datasets.}
        \end{minipage}%
        \begin{minipage}[t]{0.45\linewidth}
        \centering
        \includegraphics[height=1.5in,width=2.5in]{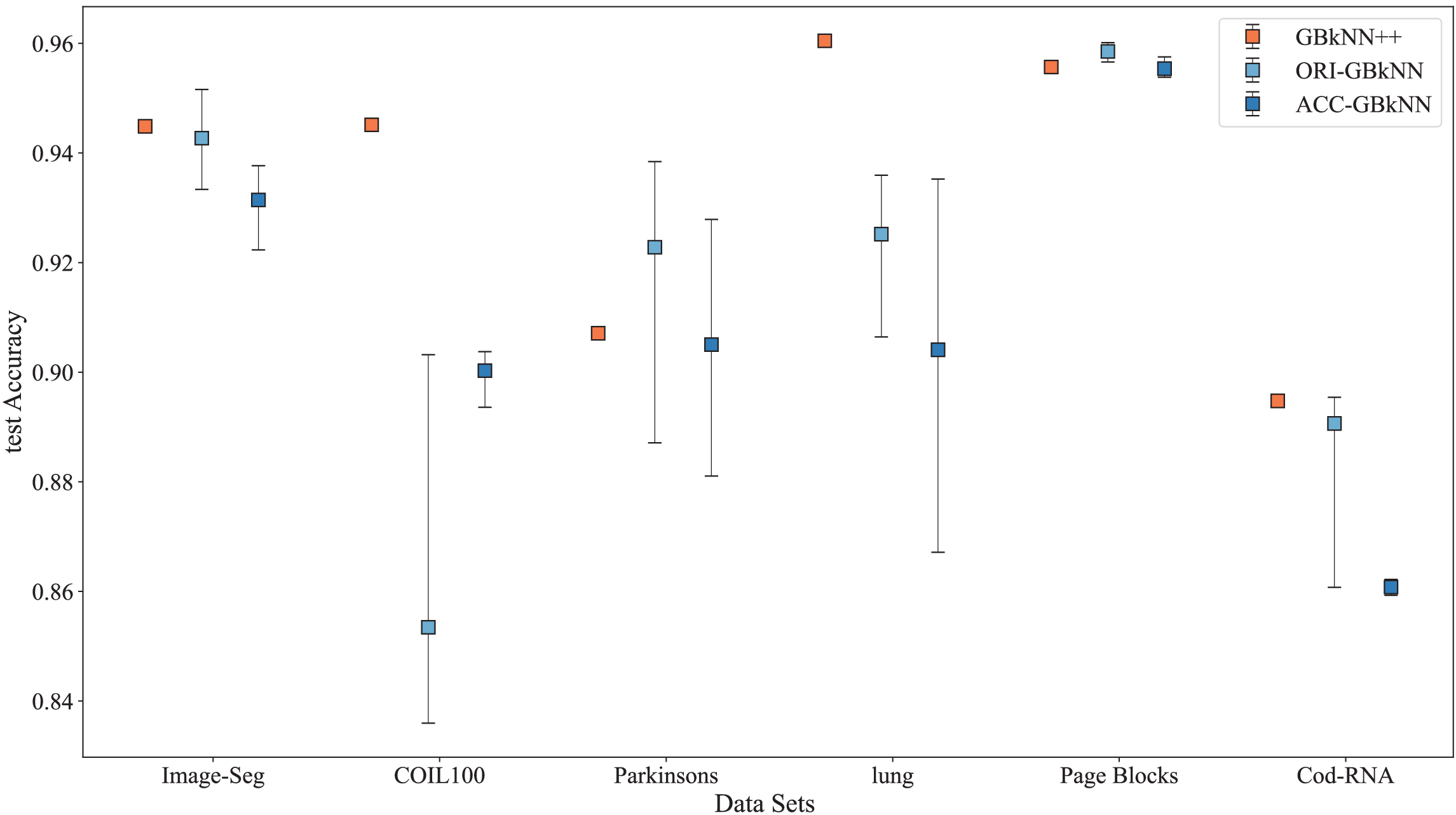}

        \small{(b) Stability on test $Accuracy$ on $2th$ group datasets.}
        \end{minipage}%
        \vspace{0.2in}
\quad
        \begin{minipage}[t]{0.45\linewidth}
        \centering
        \includegraphics[height=1.5in,width=2.5in]{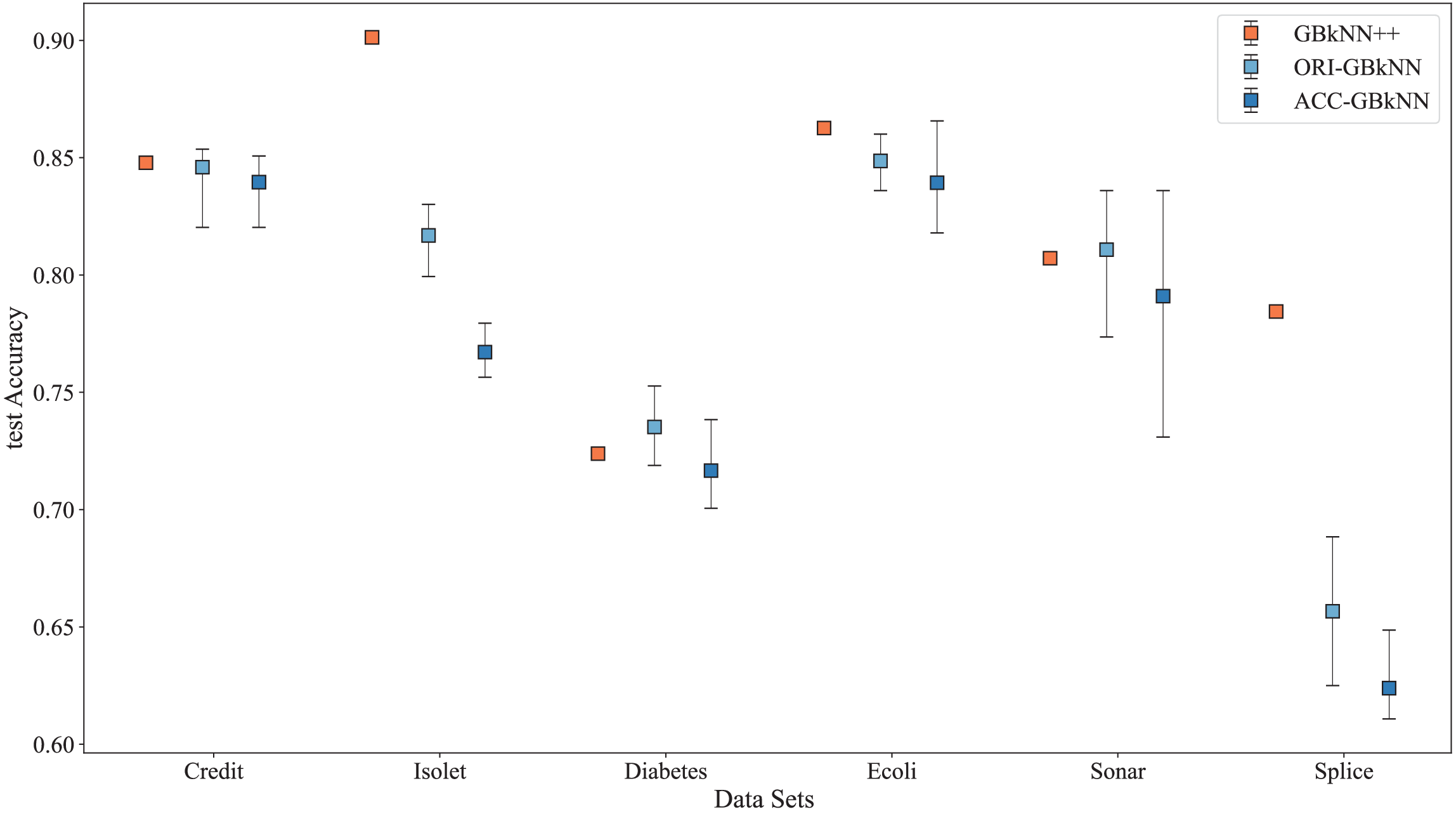}

        \small{(c) Stability on test $Accuracy$ on $3th$ group datasets.}
        \end{minipage}%
        \begin{minipage}[t]{0.45\linewidth}
        \centering
        \includegraphics[height=1.5in,width=2.5in]{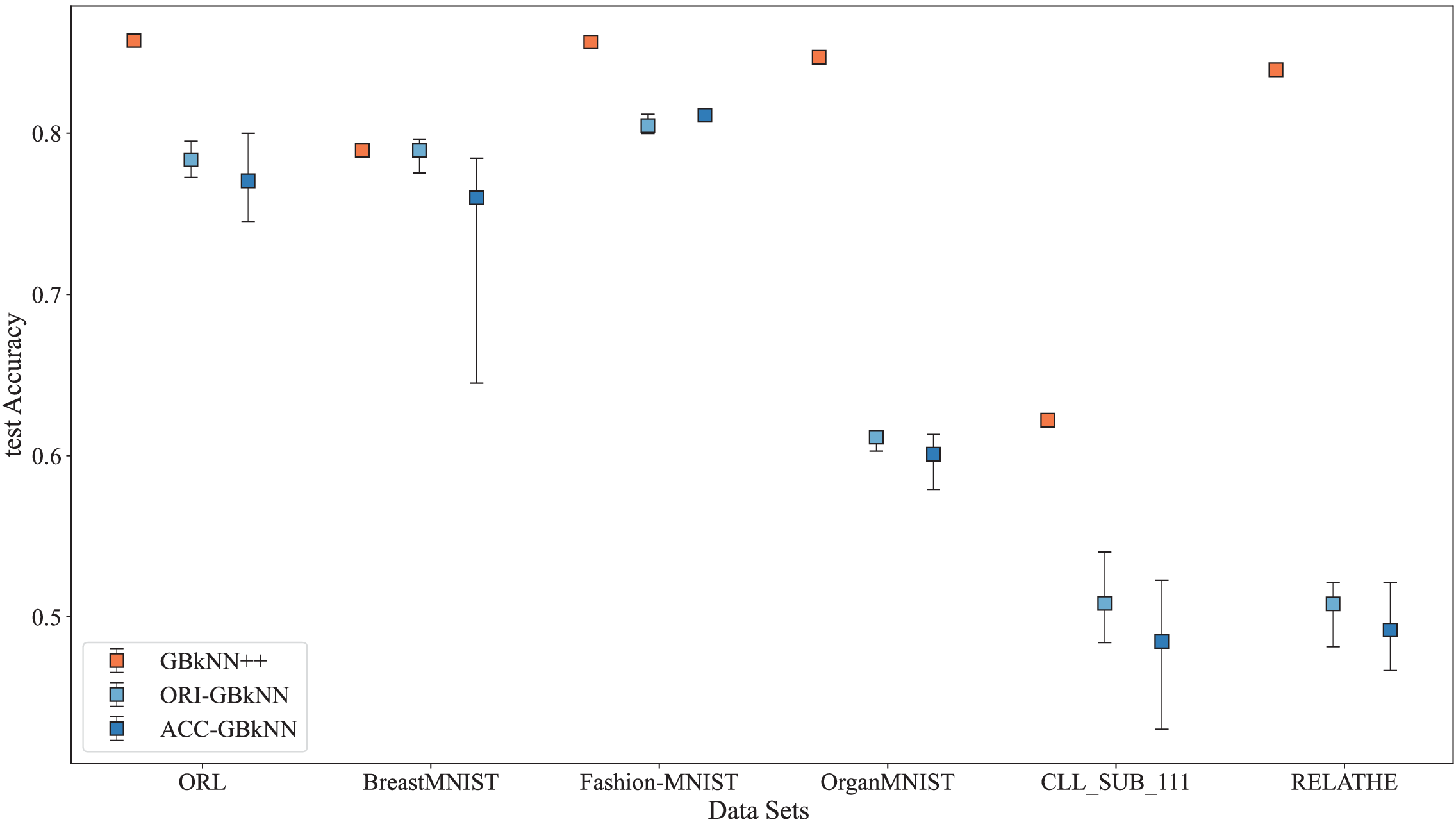}

        \small{(d) Stability on test $Accuracy$ on $4th$ group datasets.}
        \end{minipage}%

  \caption {Comparison on stability of GB-based classifiers.}
\label{fig9}

\end{figure*}

\subsection{Effectiveness}
\label{subsec:Effectiveness}
For the purpose of demonstrating the effectiveness of the proposed method, the 10-fold cross-validation is employed on GB$k$NN++ and all comparison methods on each dataset for one time. The performance of all methods is evaluated using the test $Accuracy$ and its $Sd$. The details of the experimental results are shown in Table \ref{table2}.

\begin{table*}[htbp]

	\renewcommand{\arraystretch}{1.0}
	\caption{Comparison on test $Accuracy$.}
	\label{table2}
	\centering
	\renewcommand\tabcolsep{3.5pt}
\begin{tabular}{p{2.7cm}p{0.5cm}p{1.1cm}p{0.5cm}p{1.1cm}p{0.5cm}p{1.1cm}p{0.5cm}p{1.1cm}p{0.5cm}p{1.1cm}p{0.5cm}p{1.1cm}}
	\toprule[1pt]
\multirow{2}{*}{Datasets}	&\multicolumn{2}{c}{GB$k$NN++} & \multicolumn{2}{c}{ORI-GB$k$NN} &\multicolumn{2}{c}{ACC-GB$k$NN} &\multicolumn{2}{c}{$k$NN} &\multicolumn{2}{c}{CART} &\multicolumn{2}{c}{SVM}
\\ \cmidrule(lr){2-3} \cmidrule(lr){4-5} \cmidrule(lr){6-7} \cmidrule(lr){8-9} \cmidrule(lr){10-11} \cmidrule(lr){12-13}
	 &Mean &$\pm{Sd}$	&Mean &$\pm{Sd}$ 	&Mean &$\pm{Sd}$   &Mean &$\pm{Sd}$  &Mean &$\pm{Sd}$ &Mean &$\pm{Sd}$
\\ \hline
CLL\_SUB\_111	&0.622 &$\pm{0.061}$ &0.530 &$\pm{0.148}$ &0.478 &$\pm{0.126}$	&0.581 &$\pm{0.140}$	&\textbf{0.639} &$\pm{0.142}$	&0.467 &$\pm{0.039}$	\\
Parkinsons        &0.907         &$\pm{0.061}$   &\textbf{0.934}&$\pm{0.039}$  &0.908&$\pm{0.039}$  &0.894          &$\pm{0.040}$  &0.857         &$\pm{0.068}$  &0.754        &$\pm{0.019}$  \\
lung	        &\textbf{0.960} &$\pm{0.053}$ &0.926 &$\pm{0.061}$ &0.887 &$\pm{0.026}$	&0.946 &$\pm{0.012}$	&0.907 &$\pm{0.021}$	&0.774 &$\pm{0.002}$	\\
Sonar              &$\textbf{0.807}$  &$\pm{0.095}$   &0.803          &$\pm{0.073}$  &0.775&$\pm{0.058}$  &0.765          &$\pm{0.060}$  &0.696         &$\pm{0.088}$  &0.716     &$\pm{0.084}$ \\
Ecoli              &\textbf{0.863}  &$\pm{0.049}$   &0.854          &$\pm{0.043}$  &0.836&$\pm{0.038}$  &0.850          &$\pm{0.021}$  &0.777         &$\pm{0.082}$  &0.768        &$\pm{0.026}$ \\
ORL                 &\textbf{0.857}  &$\pm{0.056}$   &0.785          &$\pm{0.061}$  &0.770&$\pm{0.045}$  &0.809          &$\pm{0.093}$  &0.617         &$\pm{0.082}$  &0.830        &$\pm{0.047}$ \\
Credit    &0.848         &$\pm{0.036}$   &0.848          &$\pm{0.040}$  &0.835&$\pm{0.043}$  &0.853          &$\pm{0.014}$  &0.806         &$\pm{0.038}$  &\textbf{0.855} &$\pm{0.024}$ \\
Diabetes            &0.724           &$\pm{0.041}$   &0.725           &$\pm{0.042}$  &0.714 &$\pm{0.051}$ &0.738         &$\pm{0.011}$  &0.702     &$\pm{0.027}$  &\textbf{0.759} &$\pm{0.052}$   \\
BreastMNIST         &\textbf{0.789}  &$\pm{0.030}$   &0.777       &$\pm{0.032}$  &0.775 &$\pm{0.029}$ &0.784       &$\pm{0.007}$  &0.721          &$\pm{0.057}$  &0.730          &$\pm{0.001}$   \\
fourclass           &0.992         &$\pm{0.009}$   &0.993        &$\pm{0.011}$  &0.986 &$\pm{0.011}$ &\textbf{0.998}  &$\pm{0.001}$  &0.987       &$\pm{0.021}$  &0.797          &$\pm{0.029}$   \\
splice              &0.784        &$\pm{0.027}$   &0.659           &$\pm{0.072}$  &0.625 &$\pm{0.051}$ &0.697         &$\pm{0.009}$  &\textbf{0.892} &$\pm{0.026}$  &0.820        &$\pm{0.043}$   \\
RELATHE	        &0.839 &$\pm{0.044}$ &0.500 &$\pm{0.050}$ &0.495 &$\pm{0.043}$	&0.780 &$\pm{0.038}$	&\textbf{0.867} &$\pm{0.028}$	&0.546 &$\pm{0.022}$	\\
COIL20              &\textbf{0.982}  &$\pm{0.010}$   &0.960           &$\pm{0.024}$  &0.945 &$\pm{0.015}$ &0.958     &$\pm{0.028}$  &0.923        &$\pm{0.017}$  &0.883        &$\pm{0.020}$   \\
Isolet	        &\textbf{0.901} &$\pm{0.031}$ &0.827 &$\pm{0.022}$ &0.781 &$\pm{0.020}$	&0.884 &$\pm{0.027}$	&0.792 &$\pm{0.030}$	&\textbf{0.901} &$\pm{0.018}$	\\
Image-Seg  &0.945           &$\pm{0.010}$   &0.939           &$\pm{0.016}$  &0.935 &$\pm{0.018}$ &\textbf{0.948}  &$\pm{0.006}$  &0.940          &$\pm{0.012}$  &0.877          &$\pm{0.032}$   \\
Page         &0.956           &$\pm{0.006}$   &0.957           &$\pm{0.006}$  &0.955 &$\pm{0.009}$ &0.955         &$\pm{0.005}$  &\textbf{0.960} &$\pm{0.006}$  &0.926          &$\pm{0.006}$   \\
svmguide1           &0.960           &$\pm{0.008}$   &0.960        &$\pm{0.010}$  &0.955 &$\pm{0.008}$ &\textbf{0.963}  &$\pm{0.004}$  &0.959        &$\pm{0.010}$  &0.954       &$\pm{0.007}$   \\
COIL100             &\textbf{0.945}  &$\pm{0.006}$   &0.848         &$\pm{0.011}$  &0.895 &$\pm{0.019}$ &0.912       &$\pm{0.049}$  &0.804      &$\pm{0.016}$  &0.684          &$\pm{0.020}$   \\
Pen                 &0.989           &$\pm{0.003}$   &0.988        &$\pm{0.002}$  &0.986 &$\pm{0.003}$ &\textbf{0.991}  &$\pm{0.002}$  &0.961        &$\pm{0.006}$  &0.978       &$\pm{0.005}$   \\
OrganMNIST          &\textbf{0.847}  &$\pm{0.007}$   &0.611        &$\pm{0.020}$  &0.601 &$\pm{0.013}$ &0.699      &$\pm{0.107}$  &0.627      &$\pm{0.008}$  &0.632          &$\pm{0.004}$   \\
Cod-RNA              &0.895           &$\pm{0.004}$   &0.894     &$\pm{0.005}$  &0.861 &$\pm{0.004}$ &0.918         &$\pm{0.021}$  &\textbf{0.949} &$\pm{0.004}$  &0.943          &$\pm{0.004}$   \\
MNIST               &0.964           &$\pm{0.003}$   &0.935       &$\pm{0.002}$  &0.955 &$\pm{0.006}$ &\textbf{0.967}  &$\pm{0.002}$  &0.876       &$\pm{0.004}$  &0.941          &$\pm{0.003}$   \\
Fashion-MNIST       &\textbf{0.857}  &$\pm{0.005}$   &0.806        &$\pm{0.004}$  &0.811 &$\pm{0.008}$ &0.855       &$\pm{0.002}$  &0.808          &$\pm{0.004}$  &\textbf{0.857} &$\pm{0.003}$   \\
Skin-Seg   & 0.995          &$\pm{0.002}$   &0.998        &$\pm{0.001}$  &0.996 &$\pm{0.001}$ &0.998     &$\pm{0.000}$  &\textbf{0.999} &$\pm{0.001}$  &0.970       &$\pm{0.002}$   \\
\hline
Mean            &\textbf{0.885}  &$\pm{0.029}$   &0.836     &$\pm{0.031}$  &0.823 &$\pm{0.029}$  &0.864          &$\pm{0.070}$  &0.836  &$\pm{0.033}$  &0.807   &$\pm{0.021}$   \\
\toprule[1pt]
\end{tabular}
\vspace{-0.5cm}
\end{table*}

In Table \ref{table2}, columns 2-7 respectively represent the test $Accuracy$ and its corresponding $Sd$ of GB$k$NN++ and all comparison methods on each dataset. The last row represents the mean test $Accuracy$ and mean $Sd$ on all datasets for each classifier. From a statistical perspective, as shown in Table \ref{table2}, the number of the datasets with the best score on test $Accuracy$ of GB$k$NN++ is respectively far higher than that of the other baselines. Furthermore, the GB$k$NN++ is the best one on mean test $Accuracy$ of all datasets. As for the $Sd$, GB$k$NN++ is at the forefront of these 6 classifiers.
In addition, GB$k$NN++ is obviously superior to the other two GB-based classifiers on most datasets.

\begin{table}[htbp]
	\renewcommand{\arraystretch}{1.0}
	\caption{Wilcoxon signed-rank test of GB$k$NN++.}
	\label{table3}
	\centering
	\renewcommand\tabcolsep{3.5pt}
\begin{tabular}{ccccccc}
	\toprule[1pt]
{Classifiers}&{ORI-GB$k$NN} &{ACC-GB$k$NN} &{$k$NN} &{CART} &{SVM}
\\ \hline
$R^+$            &260     &296    &227    &248 	&260.5 	   \\
$R^-$            &40      &4      &73     &52 	&39.5 	   \\
T                &40      &4      &73     &52 	&39.5 	   \\
\toprule[1pt]
\end{tabular}
\vspace{-0.5cm}
\end{table}

The results of the Wilcoxon rank-sum test\cite{51} of the GB$k$NN++ in terms of test $Accuracy$ compared to the comparison methods are shown in Table \ref{table3}. And according to the Wilcoxon signed-rank test critical values table, for $\alpha = 0.05$ and $N=24$, all comparison methods satisfy $T \le 91$. Thus, the null hypothesis is rejected. In other words, the proposed classifier is significantly different from comparison methods.

The reasons for the superior performance of the proposed method in effectiveness are as follows. On the one hand, there are two reasons that GB-based classifiers are superior to the other 3 classical machine learning classifiers. First, GB-based classifiers are naturally robust to noisy data. Second, since the geometry used to represent IGs is a ball, the IGs constructed based on the GBG method are more suitable for describing datasets with spherical distribution including irregular spherical distribution. On the other hand, there are two reasons why GB$k$NN++ outperforms the other two GB-based classifiers. First, the GBG++ method is based on the strategy of `priority for majority class', then the constructed GB contains more samples; namely, the support degree of the classification rule formed by the GB is greater. Thus, the quality of classification rules based on GBs constructed by GB$k$NN++ is higher. Second, the classification rules of GB$k$NN++ additionally consider the number of samples contained in the GB; thus, it performs better at class boundaries. In detail, for the queried sample, if its distances to multiple GBs are the same, the GB$k$NN++ can effectively classify it into the GB containing more samples rather than randomly dividing it into any equidistant GB. Consequently, the proposed GB$k$NN++ is more prominent in terms of effectiveness.

\vspace{-0.3cm}

\subsection{Efficiency}
\label{subsec:Efficiency}

\begin{figure}[htbp]

    \begin{minipage}[t]{0.9\linewidth}
           \centering
           \includegraphics[height=1.5in, width=3.0in]{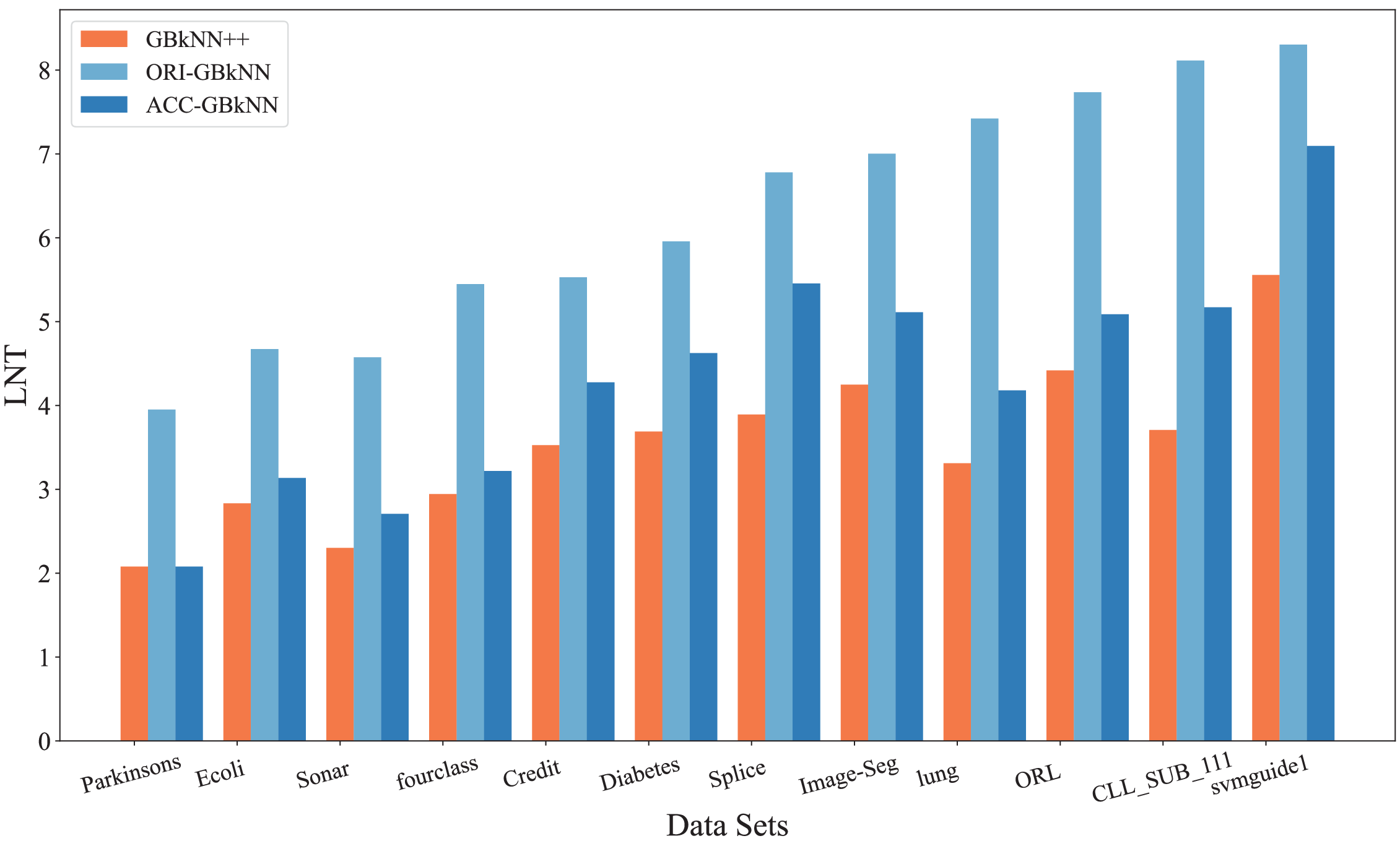}
            \small{(a) Execution time on $1th$ group datasets.}
            \label{fig:a}
    \end{minipage}

    \begin{minipage}[t]{0.9\linewidth}
            \centering
            \includegraphics[height=1.5in, width=3.0in]{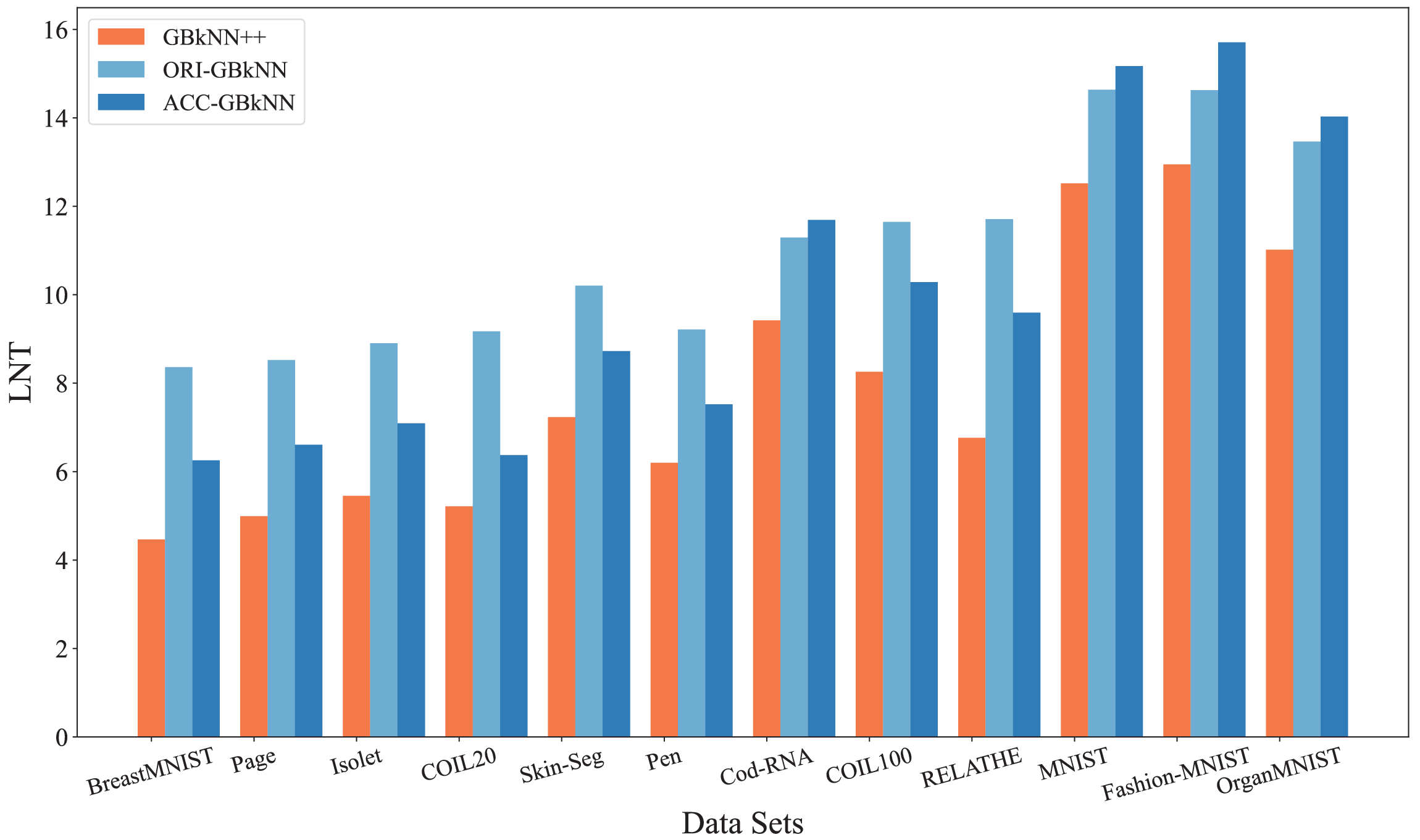}
            \small{(b) Execution time on $2th$ group datasets.}
            \label{fig:b}
    \end{minipage}
    \caption{Execution time of GB-based classifiers.}
    \vspace{-0.3cm}
    \label{fig10}
\end{figure}

For the GB-based classifiers, the logarithm of mean execution time $LNT$ of each classifier on each dataset is calculated respectively using Eq.\ref{eq13}, which are shown in Fig. \ref{fig10}.
According to Fig. \ref{fig10}, the time cost of GB$k$NN++ for each dataset is lower than that of the ORI-GB$k$NN and ACC-GB$k$NN. Specifically, for small-scale datasets such as Sonar\cite{38}, the time cost of GB$k$NN++ is several times lower than that of the others. And, the time cost of GB$k$NN++ is dozens of times lower than that of the others for large-scale or high-dimensional datasets, such as COIL100\cite{39}. When the sizes of the datasets are similar, the efficiency improvement of the proposed method is more obvious on high-dimensional datasets (such as lung\cite{57}) than on low-dimensional datasets (such as Sonar).

The reason why GB$k$NN++ is so excellent in efficiency is that, as Section \ref{subsec:split} introduced, it only needs to calculate the distances from the undivided homogenous samples to the center when splitting each GB. And, until GB reaches the granulation stopping condition, the GB's center will be adjusted according to Definition \ref{mydef4}. However, for the ORI-GB$k$NN, as introduced in Step 1 of Fig.\ref{fig2}, since it calls $k$-means when splitting each GB, it needs to calculate the distances from all samples to these $k$ centers, and it needs to iterate many times to adjust the randomly selected centers until they reach a stable state. Moreover, for the ACC-GB$k$NN, as introduced in Step 1 of Fig. \ref{fig3}, although $k$-means is replaced with $k$-division to avoid multiple iterations to adjust the random centers, it needs to calculate the distances from all samples to the $k-1$ centers. Meanwhile, as introduced in Step 4, because of the randomness of centers, the ACC-GB$k$NN needs a global division on the whole dataset to adjust the centers after all GBs are generated. Thus, the efficiency of ACC-GB$k$NN decreases rapidly as the number of GBs increases. In summary, the proposed GB$k$NN++ indeed outperforms ORI-GB$k$NN and ACC-GB$k$NN in efficiency.

\vspace{-0.2cm}

\subsection{Stability}
\label{subsec:Stability}
To verify the consistency of the classifier's multiple classification results on the same dataset, 10-fold cross-validation is employed another nine times for the GB-based classifiers. The mean, minimum, and maximum of test $Accuracy$ jointly evaluate the stability of the classification results of the classifier. Besides, the data division remains unchanged in this section. The error bar is used to visualize experimental results; that is, the central bar, the top and bottom line of the error bar, represent the average, maximum, and minimum values of the test $Accuracy$ of 10-time 10-fold cross-validation, respectively.
According to Fig. \ref{fig9}, the mean, minimum, and maximum of test $Accuracy$ obtained by GB$k$NN++ on each dataset are the same. Additionally, the classification results of ORI-GB$k$NN and ACC-GB$k$NN are inconsistent when executed multiple times on the same dataset. Consequently, the classification results of GB$k$NN++ on the same dataset are stable, but those of ORI-GB$k$NN and ACC-GB$k$NN are unstable.

The reason that the GB$k$NN++ is stable is that it does not require random parameters when the purity threshold is given. Conversely, the essence of instability of the ORI-GB$k$NN and ACC-GB$k$NN is that the random centers are required when splitting the GB. In other words, their stabilities still need improvement. Thus, the proposed method is more applicable in some scenarios where stable results are required.

\subsection{Robustness}
\label{subsec:Robustness}

\begin{figure*}[htbp]
\vspace{-0.5cm}
\centering
    \begin{minipage}[t]{0.45\linewidth}
           \centering
           \includegraphics[height=1.1in, width=3.0in, frame]{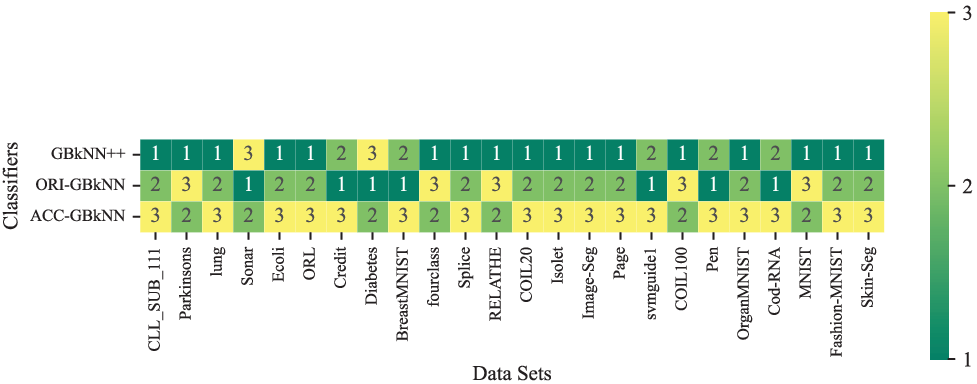}

            \small{(a) Noise rate: 10$\%$.}
            \label{fig:a}
    \end{minipage}
    \begin{minipage}[t]{0.45\linewidth}
            \centering
            \includegraphics[height=1.1in, width=3.0in, frame]{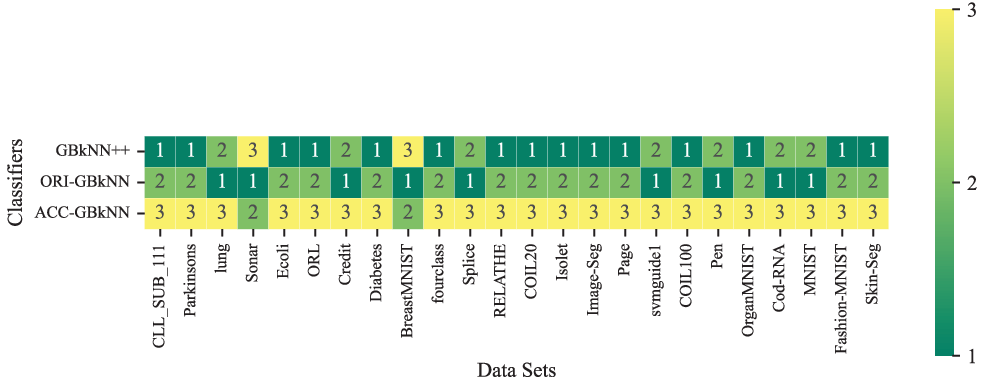}

            \small{(b) Noise rate: 20$\%$.}
            \label{fig:b}
    \vspace{0.2cm}
    \end{minipage}
    \begin{minipage}[t]{0.45\linewidth}
           \centering
           \includegraphics[height=1.1in, width=3.0in, frame]{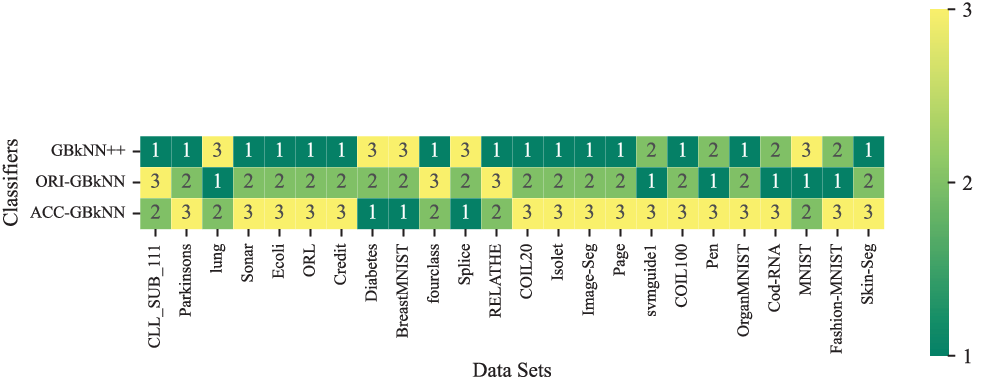}

            \small{(c) Noise rate: 30$\%$.}
            \label{fig:c}
    \end{minipage}
    \begin{minipage}[t]{0.45\linewidth}
            \centering
            \includegraphics[height=1.1in, width=3.0in, frame]{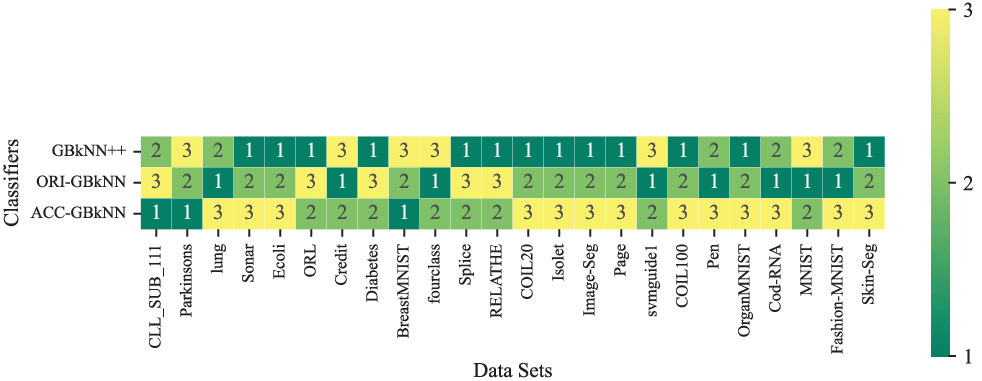}

            \small{(d) Noise rate: 40$\%$.}
            \label{fig:d}
    \vspace{0.2cm}
    \end{minipage}
\caption{Comparison on ranking of test $Accuracy$ under each noise rate.}
\label{fig11}
\vspace{-0.5cm}
\end{figure*}

To verify the robustness of the proposed method, label noise datasets with sample proportions of $10\%$, $20\%$, $30\%$, and $40\%$ are randomly constructed manually on all datasets listed in Table \ref{table1}. Specifically, samples are selected randomly on each dataset, and the labels of these samples are changed. The 10-fold cross-validation is employed on GB-based classifiers respectively on the datasets with different noise ratios, and the classifiers' robustness performance is measured by test $Accuracy$. The heat map is used to visualize the ranking of the test $Accuracy$ of the classifier on any dataset under any label noise ratio. For example, on the Parkinsons\cite{38} with label noise ratio of $10\%$, the test $Accuracy$ of GB$k$NN++ is the highest; it is marked as 1 in Fig. \ref{fig11}(a). As shown in Fig. \ref{fig11}, the GB$k$NN++ ranks first on most datasets. That is, the GB$k$NN++ outperforms ORI-GB$k$NN and ACC-GB$k$NN on test $Accuracy$ on the majority of datasets with label noise ratios of $10\%$, $20\%$, $30\%$ and $40\%$.

There are two reasons why GB$k$NN++ exhibits outstanding noise robustness. First, since the label of a GB is determined by the majority of samples, the GBG++ method naturally eliminates the influence of some noisy sample. Second, the local outlier detection is introduced into the GBG++ method when splitting each GB, which can further eliminate the outliers at the class boundary to a certain extent.

\vspace{-0.2cm}

\subsection{Parameter Sensitivity Analysis}
\label{subsec:PSA}

\begin{figure}[htbp]
\vspace{-0.1cm}
\centering
\includegraphics[height=1.6in,width=3.3in]{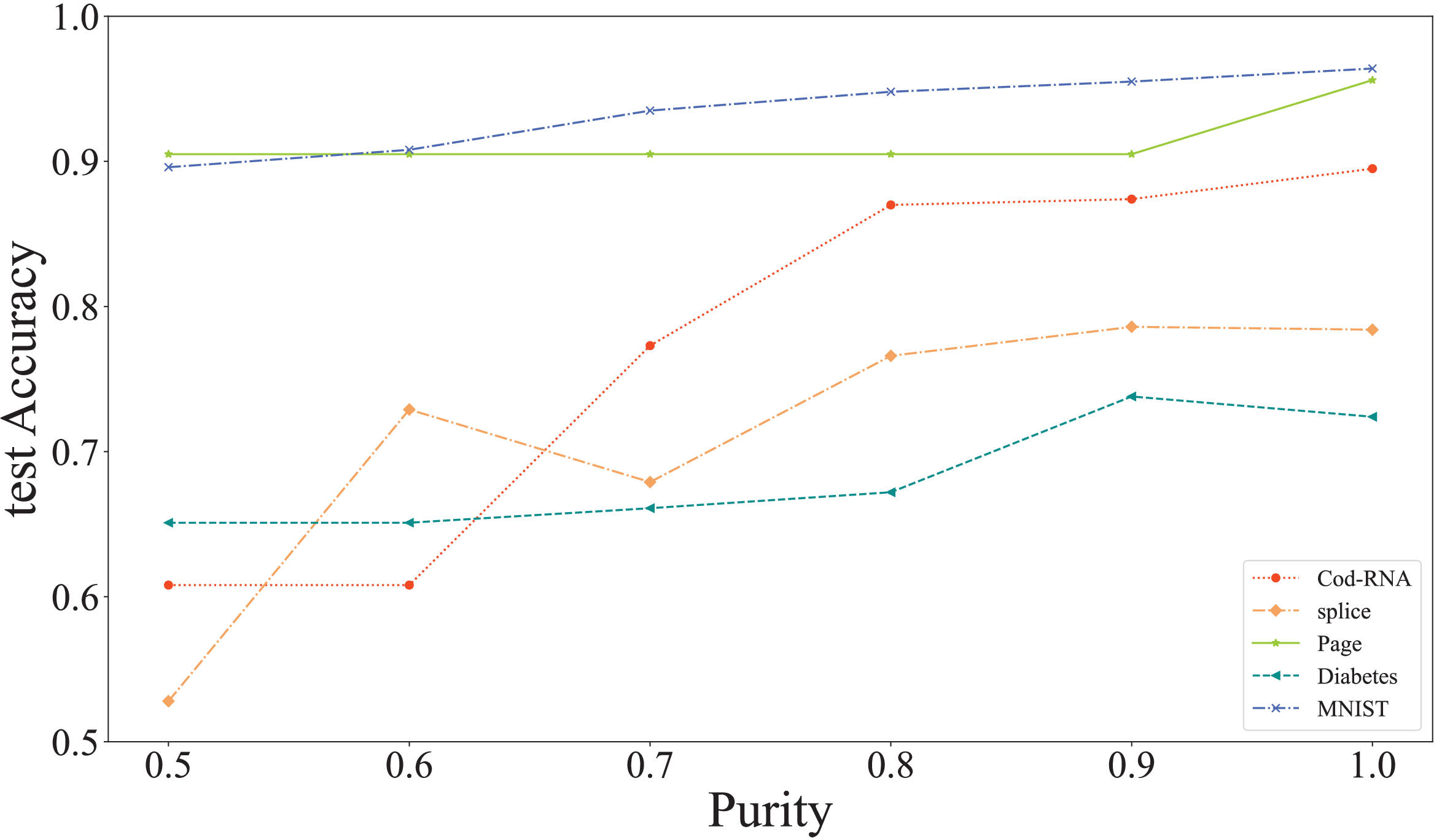}
\caption {Changes in test $Accuracy$ as the purity increases.}
\label{fig12}
\vspace{-0.3cm}
\end{figure}

Parameter sensitivity analysis is provided on the only hyperparameter in the proposed GBG++ and GB$k$NN++, which is the purity threshold. Consider 5 datasets, Cod-RNA, splice, Page, Diabetes, and MNIST, as examples to demonstrate the change of test $Accuracy$ with different parameters shown in Fig. \ref{fig12}, where threshold $P$ ranges from 0.5 to 1.0 with 0.1 as the step size. According to Fig. \ref{fig12}, the test $Accuracy$ is sensitive to the threshold when it is less than or equal to $0.8$, but when the threshold is greater than or equal to $0.9$, it is roughly insensitive. And, when the threshold is set to $1.0$, the test $Accuracy$ of most datasets reaches the optimal or second-best level.
Similar patterns can be observable in the datasets listed in Table \ref{table1}. Thus, the purity threshold for GB-based classifiers is set to $1.0$ in this paper.

The reason for this purity threshold pattern in classification tasks is that the greater the purity of the GB, the more accurate it is to use these GBs to depict the distribution of the original dataset, especially the demarcation of class boundaries.

\subsection{Ablation Study}
\label{subsec:Ablation}
\subsubsection{The Impact of AM Module}
In the proposed classifier, the AM is used to determine the center and radius of the GB on the majority class, which is a data-driven approach guided by the human cognitive pattern `global topology precedence'. Based on AM, the quality of the constructed GBs is better. According to Table \ref{table2}, the effectiveness of the GB$k$NN++ has been verified. To further demonstrate the performance improvement brought by adding AM to the GBG++, an ablation study is designed for GB$k$NN++ with and without AM. Specifically, for GB$k$NN++ without AM, when splitting the GB, it is constructed directly on all undivided samples according to Definition \ref{mydef4}. Meanwhile, the other modules remain unchanged.

As shown in Table \ref{table4}, GB$k$NN++ with AM outperforms without AM in terms of test $Accuracy$ on any dataset. Thus, the conclusion can be drawn that the introduction of AM is valuable for GBG++.

\begin{table}[htbp]
\begin{center}
	\renewcommand{\arraystretch}{1.0}
    \caption{Impact of AM module on GB$k$NN++ in terms of test $Accuracy$.}
    \label{table4}
	\centering
	\renewcommand\tabcolsep{3.6pt}
	\begin{tabular}{p{2.5cm}p{2.0cm}p{2.0cm}}
		\toprule[1pt]
    \multirow{2}{*}{Datasets}	&\multicolumn{2}{c}{GB$k$NN++}
    \\ \cmidrule(lr){2-3}
    &{with AM} & {without AM}
        \\ \hline
        Parkinsons        &\textbf{0.907}      &0.877   \\
        Sonar             &\textbf{0.807}      &0.759   \\
        Ecoli             &\textbf{0.863}      &0.827   \\
        ORL               &\textbf{0.857}      &0.393   \\
        BreastMNIST       &\textbf{0.789}      &0.768   \\
        fourclass         &\textbf{0.992}      &0.985   \\
        splice            &\textbf{0.784}      &0.727   \\
        COIL20            &\textbf{0.982}      &0.923   \\
        Image-Seg         &\textbf{0.945}      &0.927   \\
        Fashion-MNIST     &\textbf{0.857}      &0.841   \\
		\toprule[1pt]
	\end{tabular}
\vspace{-0.5cm}
\end{center}
\end{table}

\subsubsection{The Impact of Outlier Detection Module}
Moreover, another ablation study on GB$k$NN++ is conducted to evaluate the impact of the proposed outlier detection module. Based on noisy datasets with label noise ratios of $10\%$, $20\%$, $30\%$, and $40\%$ corresponding to the 10 datasets constructed in Section \ref{subsec:Robustness}, the GB$k$NN++ with and without noise detection module are trained. The performance is evaluated using test $Accuracy$. As shown in Table \ref{table5}, GB$k$NN++ with the outlier detection module (denoted as OD) outperforms GB$k$NN++ without the outlier detection module (denoted as NOD) at all label noise ratios.

In summary, the outlier detection module effectively filters out outliers, thereby improving the performance of GB$k$NN++. This module is practical because it identifies orphan GBs as outliers and drops them.

\begin{table}[htbp]
    \fontsize{7.5}{12}\selectfont
	\caption{Impact of outlier detection module on GB$k$NN++ in terms of test $Accuracy$.}
    \label{table5}
	\centering
    \setlength{\tabcolsep}{4pt}
\begin{tabular}{p{1.3cm}p{0.6cm}p{0.6cm}p{0.6cm}p{0.6cm}p{0.6cm}p{0.6cm}p{0.6cm}p{0.6cm}}
	\toprule[1pt]
\multirow{2}{*}{Datasets}	&\multicolumn{2}{@{\hspace{-1em}}c}{Noise rate:10$\%$} & \multicolumn{2}{@{\hspace{-1em}}c}{Noise rate:20$\%$} &\multicolumn{2}{@{\hspace{-1em}}c}{Noise rate:30$\%$} &\multicolumn{2}{@{\hspace{-1em}}c}{Noise rate:40$\%$}
\\ \cmidrule(lr){2-3} \cmidrule(lr){4-5} \cmidrule(lr){6-7} \cmidrule(lr){8-9}
	 &OD &NOD	&OD &NOD 	&OD &NOD   &OD &NOD
\\ \hline
Credit       &\textbf{0.759} 	&0.690    &\textbf{0.639} 	&0.580     &\textbf{0.562} 	&0.549    &\textbf{0.475} 	&0.464   \\
Diabetes     &\textbf{0.615} 	&0.596    &\textbf{0.581} 	&0.551     &\textbf{0.520} 	&0.498    &\textbf{0.535} 	&\textbf{0.535}   \\
Page         &\textbf{0.849} 	&0.720    &\textbf{0.734} 	&0.569     &\textbf{0.615} 	&0.457    &\textbf{0.481} 	&0.353   \\
svmguide1    &\textbf{0.843} 	&0.746    &\textbf{0.719} 	&0.634     &\textbf{0.613} 	&0.572    &\textbf{0.523} 	&0.512   \\
COIL100      &\textbf{0.840} 	&0.751    &\textbf{0.731} 	&0.578     &\textbf{0.622} 	&0.434    &\textbf{0.501} 	&0.312   \\
Pen          &\textbf{0.885} 	&0.734    &\textbf{0.769} 	&0.565     &\textbf{0.644} 	&0.440    &\textbf{0.508} 	&0.325   \\
OrganMNIST   &\textbf{0.739} 	&0.712    &\textbf{0.624} 	&0.569     &\textbf{0.504} 	&0.448    &\textbf{0.394} 	&0.347   \\
Cod-RNA      &\textbf{0.787} 	&0.711    &\textbf{0.679} 	&0.617     &\textbf{0.583} 	&0.548    &\textbf{0.523} 	&0.517   \\
MNIST        &\textbf{0.857} 	&0.788    &\textbf{0.738} 	&0.641     &\textbf{0.606} 	&0.510    &\textbf{0.475} 	&0.393   \\
Skin-Seg     &\textbf{0.875} 	&0.764    &\textbf{0.741} 	&0.642     &\textbf{0.617} 	&0.563    &\textbf{0.530} 	&0.518   \\
\toprule[1pt]
\end{tabular}
\vspace{-0.5cm}
\end{table}

\section{Conclusion}
\label{sec:Conclusion}
\noindent GBC is an efficient, robust, and scalable GrC method. This paper reviews the original GBG method, an acceleration GBG method, and the original GB$k$NN. Considering the shortcomings of the existing GBC, this paper proposes a new GBG method (i.e., GBG++) and an improved GB$k$NN (i.e., GB$k$NN++). The experimental results show that GBG++ is stable because it does not require any random center. The GBG++ is efficient because it only needs to calculate the distance from undivided samples to a data-driven center that is calculated based on the majority class of these undivided samples when splitting the GB. The GBG++ is more robust because it detects local outliers when splitting the GB. Furthermore, the GB$k$NN++ is more effective because it tackles the problem that only using the geometric characteristics of GB fails to describe its actual quality. The results of the comparative experiments verify that the proposed GB$k$NN++ outperforms other comparison methods, including $2$ state-of-the-art GB-based classifiers and $3$ classical machine learning classifiers, on $24$ public benchmark datasets.

Moreover, deep learning has made significant progress in various fields. In future works, GBC will be introduced into the artificial neural networks. Specifically, the research tasks include optimizing the input space in a task-driven manner and granulating the network structure using the idea of multi-granularity to tackle the problems of complex structure, time-consuming nature, and inexplicability for deep neural networks.

\ifCLASSOPTIONcaptionsoff
  \newpage
\fi

\bibliography{mybibfile}

\begin{thebibliography}{10}
\providecommand{\url}[1]{#1}
\csname url@samestyle\endcsname
\providecommand{\newblock}{\relax}
\providecommand{\bibinfo}[2]{#2}
\providecommand{\BIBentrySTDinterwordspacing}{\spaceskip=0pt\relax}
\providecommand{\BIBentryALTinterwordstretchfactor}{4}
\providecommand{\BIBentryALTinterwordspacing}{\spaceskip=\fontdimen2\font plus
\BIBentryALTinterwordstretchfactor\fontdimen3\font minus
  \fontdimen4\font\relax}
\providecommand{\BIBforeignlanguage}[2]{{%
\expandafter\ifx\csname l@#1\endcsname\relax
\typeout{** WARNING: IEEEtranS.bst: No hyphenation pattern has been}%
\typeout{** loaded for the language `#1'. Using the pattern for}%
\typeout{** the default language instead.}%
\else
\language=\csname l@#1\endcsname
\fi
#2}}
\providecommand{\BIBdecl}{\relax}
\BIBdecl

\bibitem{5}
J.~Ba, V.~Mnih, and K.~Kavukcuoglu, ``Multiple object recognition with visual
  attention,'' \emph{arXiv preprint arXiv:1412.7755}, 2014.

\bibitem{3}
D.~Bahdanau, K.~Cho, and Y.~Bengio, ``Neural machine translation by jointly
  learning to align and translate,'' \emph{arXiv preprint arXiv:1409.0473},
  2014.

\bibitem{38}
\BIBentryALTinterwordspacing
C.~J.~M. C.~L.~Blake. Uci repository of machine learning databases. 2022, 11
  15. [Online]. Available: \url{https://archive.ics.uci.edu/ml/datasets.php}
\BIBentrySTDinterwordspacing

\bibitem{39}
D.~Cai, X.~He, J.~Han, and T.~S. Huang, ``Graph regularized nonnegative matrix
  factorization for data representation,'' \emph{IEEE transactions on pattern
  analysis and machine intelligence}, vol.~33, no.~8, pp. 1548--1560, 2010.

\bibitem{40}
D.~Cai, X.~He, J.~Han, and H.-J. Zhang, ``Orthogonal laplacianfaces for face
  recognition,'' \emph{IEEE transactions on image processing}, vol.~15, no.~11,
  pp. 3608--3614, 2006.

\bibitem{7}
L.~Chen, ``Topological structure in visual perception,'' \emph{Science}, vol.
  218, no. 4573, pp. 699--700, 1982.

\bibitem{51}
J.~Dem{\v{s}}ar, ``Statistical comparisons of classifiers over multiple data
  sets,'' \emph{The Journal of Machine learning research}, vol.~7, pp. 1--30,
  2006.

\bibitem{27}
S.~Ding, X.~Zhang, Y.~An, and Y.~Xue, ``Weighted linear loss multiple birth
  support vector machine based on information granulation for multi-class
  classification,'' \emph{Pattern Recognition}, vol.~67, pp. 32--46, 2017.

\bibitem{33}
K.~Fukunaga and P.~M. Narendra, ``A branch and bound algorithm for computing
  k-nearest neighbors,'' \emph{IEEE transactions on computers}, vol. 100,
  no.~7, pp. 750--753, 1975.

\bibitem{4}
A.~Galassi, M.~Lippi, and P.~Torroni, ``Attention in natural language
  processing,'' \emph{IEEE Transactions on Neural Networks and Learning
  Systems}, vol.~32, no.~10, pp. 4291--4308, 2020.

\bibitem{1}
W.~Guoyin and Y.~Hong, ``Multi-granularity cognitive computing—a new model
  for big data intelligent computing,'' \emph{Frontiers of Data and Domputing},
  vol.~1, no.~2, pp. 75--85, 2020.

\bibitem{45}
T.~K. Ho and E.~M. Kleinberg, ``Building projectable classifiers of arbitrary
  complexity,'' in \emph{Proceedings of 13th International Conference on
  Pattern Recognition}, vol.~2.\hskip 1em plus 0.5em minus 0.4em\relax IEEE,
  1996, pp. 880--885.

\bibitem{46}
C.-W. Hsu, C.-C. Chang, C.-J. Lin \emph{et~al.}, ``A practical guide to support
  vector classification,'' 2003.

\bibitem{35}
B.~S. Kim and S.~B. Park, ``A fast k nearest neighbor finding algorithm based
  on the ordered partition,'' \emph{IEEE Transactions on Pattern Analysis and
  Machine Intelligence}, no.~6, pp. 761--766, 1986.

\bibitem{42}
Y.~LeCun, L.~Bottou, Y.~Bengio, and P.~Haffner, ``Gradient-based learning
  applied to document recognition,'' \emph{Proceedings of the IEEE}, vol.~86,
  no.~11, pp. 2278--2324, 1998.

\bibitem{57}
J.~Li, K.~Cheng, S.~Wang, F.~Morstatter, R.~P. Trevino, J.~Tang, and H.~Liu,
  ``Feature selection: A data perspective,'' \emph{ACM computing surveys
  (CSUR)}, vol.~50, no.~6, pp. 1--45, 2017.

\bibitem{34}
Z.~Li, S.~Wang, H.~Yu, Y.~Zhu, Q.~Wu, L.~Wang, Z.~Wu, Y.~Gan, W.~Li, B.~Qiu,
  and J.~Tian, ``A novel deep learning framework based mask-guided attention
  mechanism for distant metastasis prediction of lung cancer,'' \emph{IEEE
  Transactions on Emerging Topics in Computational Intelligence}, vol.~7,
  no.~2, pp. 330--341, 2023.

\bibitem{36}
J.~McNames, ``A fast nearest-neighbor algorithm based on a principal axis
  search tree,'' \emph{IEEE Transactions on pattern analysis and machine
  intelligence}, vol.~23, no.~9, pp. 964--976, 2001.

\bibitem{2}
E.~A. Nadaraya, ``On estimating regression,'' \emph{Theory of Probability \&
  Its Applications}, vol.~9, no.~1, pp. 141--142, 1964.

\bibitem{21}
A.~Onan, ``An ensemble scheme based on language function analysis and feature
  engineering for text genre classification,'' \emph{Journal of Information
  Science}, vol.~44, no.~1, pp. 28--47, 2018.

\bibitem{28}
------, ``Bidirectional convolutional recurrent neural network architecture
  with group-wise enhancement mechanism for text sentiment classification,''
  \emph{Journal of King Saud University-Computer and Information Sciences},
  vol.~34, no.~5, pp. 2098--2117, 2022.

\bibitem{50}
A.~Onan, S.~Koruko{\u{g}}lu, and H.~Bulut, ``Ensemble of keyword extraction
  methods and classifiers in text classification,'' \emph{Expert Systems with
  Applications}, vol.~57, pp. 232--247, 2016.

\bibitem{52}
------, ``A hybrid ensemble pruning approach based on consensus clustering and
  multi-objective evolutionary algorithm for sentiment classification,''
  \emph{Information Processing \& Management}, vol.~53, no.~4, pp. 814--833,
  2017.

\bibitem{20}
T.~Ouyang, W.~Pedrycz, and N.~J. Pizzi, ``Rule-based modeling with dbscan-based
  information granules,'' \emph{IEEE Transactions on Cybernetics}, vol.~51,
  no.~7, pp. 3653--3663, 2019.

\bibitem{25}
H.-S. Park, W.~Pedrycz, and S.-K. Oh, ``Granular neural networks and their
  development through context-based clustering and adjustable dimensionality of
  receptive fields,'' \emph{IEEE transactions on neural networks}, vol.~20,
  no.~10, pp. 1604--1616, 2009.

\bibitem{12}
Z.~Pawlak, ``Rough sets,'' \emph{International journal of computer \&
  information sciences}, vol.~11, no.~5, pp. 341--356, 1982.

\bibitem{24}
W.~Pedrycz and G.~Vukovich, ``Granular neural networks,''
  \emph{Neurocomputing}, vol.~36, no. 1-4, pp. 205--224, 2001.

\bibitem{48}
A.~Pramanik, S.~K. Pal, J.~Maiti, and P.~Mitra, ``Granulated rcnn and
  multi-class deep sort for multi-object detection and tracking,'' \emph{IEEE
  Transactions on Emerging Topics in Computational Intelligence}, vol.~6,
  no.~1, pp. 171--181, 2021.

\bibitem{9}
A.~Rodriguez and A.~Laio, ``Clustering by fast search and find of density
  peaks,'' \emph{science}, vol. 344, no. 6191, pp. 1492--1496, 2014.

\bibitem{23}
S.~Salehi, A.~Selamat, M.~R. Mashinchi, and H.~Fujita, ``The synergistic
  combination of particle swarm optimization and fuzzy sets to design granular
  classifier,'' \emph{Knowledge-Based Systems}, vol.~76, pp. 200--218, 2015.

\bibitem{47}
Y.~Tang, Z.~Pan, W.~Pedrycz, F.~Ren, and X.~Song, ``Viewpoint-based kernel
  fuzzy clustering with weight information granules,'' \emph{IEEE Transactions
  on Emerging Topics in Computational Intelligence}, vol.~7, no.~2, pp.
  342--356, 2023.

\bibitem{49}
Y.~Tang, W.~Pedrycz, and F.~Ren, ``Granular symmetric implicational method,''
  \emph{IEEE Transactions on Emerging Topics in Computational Intelligence},
  vol.~6, no.~3, pp. 710--723, 2021.

\bibitem{44}
A.~V. Uzilov, J.~M. Keegan, and D.~H. Mathews, ``Detection of non-coding rnas
  on the basis of predicted secondary structure formation free energy change,''
  \emph{BMC bioinformatics}, vol.~7, no.~1, pp. 1--30, 2006.

\bibitem{10}
C.~Wang, W.~Pedrycz, Z.~Li, M.~Zhou, and S.~S. Ge, ``G-image segmentation:
  similarity-preserving fuzzy c-means with spatial information constraint in
  wavelet space,'' \emph{IEEE Transactions on Fuzzy Systems}, vol.~29, no.~12,
  pp. 3887--3898, 2020.

\bibitem{53}
H.~Wang, Y.~Cheng, C.~L.~P. Chen, and X.~Wang, ``Broad graph convolutional
  neural network and its application in hyperspectral image classification,''
  \emph{IEEE Transactions on Emerging Topics in Computational Intelligence},
  vol.~7, no.~2, pp. 610--616, 2023.

\bibitem{26}
W.~Wang, W.~Liu, and H.~Chen, ``Information granules-based bp neural network
  for long-term prediction of time series,'' \emph{IEEE Transactions on Fuzzy
  Systems}, vol.~29, no.~10, pp. 2975--2987, 2020.

\bibitem{32}
S.~Xia, X.~Dai, G.~Wang, X.~Gao, and E.~Giem, ``An efficient and adaptive
  granular-ball generation method in classification problem,'' \emph{IEEE
  Transactions on Neural Networks and Learning Systems}, pp. 1--13, 2022.

\bibitem{29}
S.~Xia, Y.~Liu, X.~Ding, G.~Wang, H.~Yu, and Y.~Luo, ``Granular ball computing
  classifiers for efficient, scalable and robust learning,'' \emph{Information
  Sciences}, vol. 483, pp. 136--152, 2019.

\bibitem{30}
S.~Xia, D.~Peng, D.~Meng, C.~Zhang, G.~Wang, E.~Giem, W.~Wei, and Z.~Chen, ``A
  fast adaptive k-means with no bounds,'' \emph{IEEE Transactions on Pattern
  Analysis and Machine Intelligence}, 2020.

\bibitem{22}
S.~Xia, G.~Wang, Z.~Chen, Y.~Duan \emph{et~al.}, ``Complete random forest based
  class noise filtering learning for improving the generalizability of
  classifiers,'' \emph{IEEE Transactions on Knowledge and Data Engineering},
  vol.~31, no.~11, pp. 2063--2078, 2018.

\bibitem{31}
S.~Xia, S.~Zheng, G.~Wang, X.~Gao, and B.~Wang, ``Granular ball sampling for
  noisy label classification or imbalanced classification,'' \emph{IEEE
  Transactions on Neural Networks and Learning Systems}, 2021.

\bibitem{43}
H.~Xiao, K.~Rasul, and R.~Vollgraf, ``Fashion-mnist: a novel image dataset for
  benchmarking machine learning algorithms,'' \emph{arXiv preprint
  arXiv:1708.07747}, 2017.

\bibitem{6}
K.~Xu, J.~Ba, R.~Kiros, K.~Cho, A.~Courville, R.~Salakhudinov, R.~Zemel, and
  Y.~Bengio, ``Show, attend and tell: Neural image caption generation with
  visual attention,'' in \emph{International conference on machine
  learning}.\hskip 1em plus 0.5em minus 0.4em\relax PMLR, 2015, pp. 2048--2057.

\bibitem{41}
J.~Yang, R.~Shi, and B.~Ni, ``Medmnist classification decathlon: A lightweight
  automl benchmark for medical image analysis,'' in \emph{2021 IEEE 18th
  International Symposium on Biomedical Imaging (ISBI)}.\hskip 1em plus 0.5em
  minus 0.4em\relax IEEE, 2021, pp. 191--195.

\bibitem{14}
Y.~Yao, ``Three-way decisions with probabilistic rough sets,''
  \emph{Information sciences}, vol. 180, no.~3, pp. 341--353, 2010.

\bibitem{11}
L.~A. Zadeh, ``Fuzzy sets,'' \emph{Information and control}, vol.~8, no.~3, pp.
  338--353, 1965.

\bibitem{8}
------, ``Fuzzy sets and information granularity,'' \emph{Fuzzy sets, fuzzy
  logic, and fuzzy systems: selected papers}, pp. 433--448, 1979.

\bibitem{19}
L.~Zhang, H.~Li, X.~Zhou, and B.~Huang, ``Sequential three-way decision based
  on multi-granular autoencoder features,'' \emph{Information sciences}, vol.
  507, pp. 630--643, 2020.

\bibitem{13}
L.~Zhang and B.~Zhang, ``The quotient space theory of problem solving,''
  \emph{Fundamenta Informaticae}, vol.~59, no. 2-3, pp. 287--298, 2004.

\bibitem{15}
Q.~Zhang, Y.~Cheng, F.~Zhao, G.~Wang, and S.~Xia, ``Optimal scale combination
  selection integrating three-way decision with hasse diagram,'' \emph{IEEE
  Transactions on Neural Networks and Learning Systems}, 2021.

\bibitem{18}
Q.~Zhang, C.~Yang, and G.~Wang, ``A sequential three-way decision model with
  intuitionistic fuzzy numbers,'' \emph{IEEE transactions on systems, man, and
  cybernetics: systems}, vol.~51, no.~5, pp. 2640--2652, 2019.

\bibitem{56}
Z.~Zhang, W.~Jiang, J.~Qin, L.~Zhang, F.~Li, M.~Zhang, and S.~Yan, ``Jointly
  learning structured analysis discriminative dictionary and analysis
  multiclass classifier,'' \emph{IEEE transactions on neural networks and
  learning systems}, vol.~29, no.~8, pp. 3798--3814, 2017.

\bibitem{55}
Z.~Zhang, F.~Li, L.~Jia, J.~Qin, L.~Zhang, and S.~Yan, ``Robust adaptive
  embedded label propagation with weight learning for inductive
  classification,'' \emph{IEEE transactions on neural networks and learning
  systems}, vol.~29, no.~8, pp. 3388--3403, 2017.

\bibitem{54}
Z.~Zhang, F.~Li, M.~Zhao, L.~Zhang, and S.~Yan, ``Robust neighborhood
  preserving projection by nuclear/l2, 1-norm regularization for image feature
  extraction,'' \emph{IEEE Transactions on Image Processing}, vol.~26, no.~4,
  pp. 1607--1622, 2017.

\end{thebibliography}

\end{document}